\documentclass{ieeeaccess}
\usepackage{amsmath}
\usepackage{multirow}
\usepackage{amsthm}
\usepackage{amssymb }
\usepackage{amsmath,tabularx}
\usepackage{bm}
\usepackage{cite}
\usepackage{float}
\usepackage{mathrsfs}
\usepackage{amsfonts}
\usepackage{subcaption}
\usepackage{graphicx}
\usepackage{algorithm, algpseudocode}
\usepackage[dvipsnames]{xcolor}
\def\aL{\mathcal{L}}

\def\mS{\mathcal{S}}
\def\mA{\mathcal{A}}
\def\MMK{\text{MM-KTD}}

\def\mP{\mathcal{P}}
\def\mR{\mathcal{R}}
\def\ua{_{a}}
\def\k{_{k}}
\def\nk{_{k+1}}
\def\i{^{i}}
\def\bt{\bm{\theta}}
\def\bmu{\bm{\mu}}
\def\h{\bm{h}}

\def\I{\bm{I}}
\def\Q{\bm{Q}}

\def\K{\bm{K}}

\def\P{\bm{P}}
\def\F{\bm{F}}

\def\X{\bm{X}}
\def\Y{\bm{Y}}

\def\W{\bm{W}}
\def\w{\bm{w}}

\def\u{\bm{\mu}}
\def\Sig{\bm{\Sigma}}

\def\s{\bm{s}}
\def\pk{_{k-1}}
\def\kpk{_{k|k-1}}

\def\x{\bm{x}}

\def\y{\bm{y}}

\def\i{^{(i)}}

\def\S{S}

\algnewcommand\Input{\item[\hspace{6pt}\textbf{Input:}]}
\algnewcommand\Output{\item[\hspace{6pt}\textbf{Output:}]}
\algnewcommand\OutputVal{\textbf{output} }
\graphicspath{{}}

\makeatletter
\newcommand{\multiline}[1]{%
  \begin{tabularx}{\dimexpr\linewidth-\ALG@thistlm}[t]{@{}X@{}}
    #1
  \end{tabularx}
}
\makeatother
\def\BibTeX{{\rm B\kern-.05em{\sc i\kern-.025em b}\kern-.08em
    T\kern-.1667em\lower.7ex\hbox{E}\kern-.125emX}}

\begin{document}
\history{}
\doi{}

\title{MM-KTD: Multiple Model Kalman Temporal Differences for Reinforcement Learning}
\author{\uppercase{Parvin Malekzadeh}\authorrefmark{1}, \IEEEmembership{Student Member, IEEE},
\uppercase{Mohammad Salimibeni}\authorrefmark{1}, \IEEEmembership{Student Member, IEEE},
\uppercase{Arash Mohammadi}\authorrefmark{1}, \IEEEmembership{Senior Member, IEEE},
\uppercase{Akbar Assa}\authorrefmark{2}, \IEEEmembership{Member, IEEE}, and
\uppercase{Konstantinos N. Plataniotis}\authorrefmark{2}, \IEEEmembership{Fellow, IEEE}}
\address[1]{Concordia Institute for Information Systems Engineering,  Concordia University, Montreal, QC, Canada H3G-2W1}
\address[2]{Department of Electrical and Computer Engineering, University of Toronto, Toronto, ON, Canada}
\tfootnote{This Project was partially supported by the Department of National Defence's Innovation for Defence Excellence and Security (IDEaS) program, Canada.\vspace{-.1in}}

\markboth
{Malekzadeh \headeretal: MM-KTD for Reinforcement Learning}
{Author \headeretal: MM-KTD for Reinforcement Learning}

\corresp{Corresponding author: Arash Mohammadi (e-mail: arash.mohammadi@concordia.ca).}

\begin{abstract}
\textbf{\textit{Background}}: There has been an increasing surge of interest on development of advanced Reinforcement Learning (RL) systems as intelligent approaches to learn optimal control policies directly from smart agents' interactions with the environment. \textbf{\textit{Objectives}}: In a model-free RL method with continuous state-space, typically, the value function of the states needs to be approximated. In this regard, Deep Neural Networks (DNNs) provide an attractive modeling mechanism to approximate the value function using sample transitions. DNN-based solutions, however, suffer from high sensitivity to parameter selection, are prone to overfitting, and are not very sample efficient. A Kalman-based methodology, on the other hand, could be used as an efficient alternative. Such an approach, however, commonly requires a-priori information about the system (such as noise statistics) to perform efficiently. The main objective of this paper is to address this issue. \textbf{\textit{Methods}}: As a remedy to the aforementioned  problems, this paper proposes an innovative Multiple Model Kalman Temporal Difference (MM-KTD) framework, which adapts the parameters of the filter using the observed states and rewards. Moreover, an active learning method is proposed to enhance the sampling efficiency of the system. More specifically, the estimated uncertainty of the value functions are exploited to form the behaviour policy leading to more visits to less certain values, therefore, improving the overall learning sample efficiency. As a result, the proposed MM-KTD framework can learn the optimal policy with significantly reduced number of samples as compared to its DNN-based counterparts. \textbf{\textit{Results}}: To evaluate performance of the proposed MM-KTD framework,  we have performed a comprehensive set of experiments based on three RL benchmarks, namely, Inverted Pendulum; Mountain Car, and; Lunar Lander.  Experimental results show superiority of the proposed MM-KTD framework in comparison to its state-of-the-art counterparts.
\vspace{-.15in}
\end{abstract}

\begin{keywords}
Reinforcement Learning, Kalman Temporal Difference, Multiple Model Adaptive Estimation, Model-free, Q-Learning.\vspace{-.15in}
\end{keywords}
\titlepgskip=-15pt

\maketitle
\section{Introduction} \label{sec:Introduction}
\PARstart{I}{nspired} by exceptional learning capabilities of human beings, Reinforcement Learning (RL) systems have emerged aiming to form  optimal control policies merely by relying on the knowledge about the past interactions of an agent with its environment. Such a learning approach is particularly beneficial, as unlike supervised learning methods, an RL system does not require availability of carefully labelled data, which is typically difficult to acquire. On the contrary, to choose the best action possible at each time step, an RL system rather relies on the reward (feedback) received from each move~\cite{Spano, Seo, DrMing1, DrMing2}. Given their phenomenal potentials, there has been an increasing surge of interest on advancement of RL systems, which is also the focus of the paper.

When an accurate model of the environment is available, methods such as dynamic programming~\cite{1} may be used to find the optimal policy. An accurate model of the environment, however, is rarely available in practice leading to the need to resort to model-based RL or model-free approaches. In the model-based RL approaches, the goal is to learn the model of the system using the past transitions of the system and then apply a model-based control scheme (e.g., model predictive controller~\cite{2}) to reach the desired goal~\cite{3}. While model-based RL approaches are proven to be sample efficient, their performance, typically, degrades over time by the uncertainties and bias in the constructed model of the environment. On the other hand, model-free approaches directly acquire the optimal policy for the system, without developing a model of the system. Due to their superior asymptotic performance, model-free approaches are commonly preferred over their model-based counterparts especially when sampling the agent's past trajectories is inexpensive. The paper, focuses on development of a model-free RL approach to achieve the desirable policy for the underlying system.

A common practice in model-free RL schemes is to calculate the state (or alternatively state-action) value function for each state. While this can be done with relative ease for a system with limited number of states (e.g., via Monte Carlo sampling~\cite{4}), it is impractical to find the value function for each state when the number of states is large or the states are continuous (due to the curse of dimensionality~\cite{5}). In such cases, which entail most realistic applications, the value function needs to be approximated. In a large group of works, e.g.,~\cite{6,Tang, Hu, Kim, Xie}, artificial neural networks were employed to approximate the value function over the entire state-space. Despite few successful trials, most early attempts of such were not very successful due to the overfitting problem. However, this problem was overcome in~\cite{7}, where simple though efficient techniques were employed to avoid the overfitting problem. This work was later extended to systems with continuous actions~\cite{8}. However, it was shown later that such approaches are highly sensitive to parameter selection, therefore, required to be revised further to be applicable to a larger variety of problems. Despite continuous attempts~\cite{9, 10, 11, 12} to overcome such problems, the training of the neural network has remained an open problem in the field.

Another well-practiced approach to approximate the value function is to employ a set of weighted local estimators and convert the approximation problem to a weight estimation problem. Various local estimator were proposed in the literature, among which Radial Basis Functions (RBFs)~\cite{13}, and Cerebellar Model Articulation Controllers (CMACs)~\cite{14} are most popular. It was shown that RBFs are more suitable than CMACs in systems with continuous states, due to their continuous nature~\cite{15}. More recently, Fourier basis was proposed as the local estimator function, however, the performance of the system was shown to be comparable to those using RBFs~\cite{16}. Due to its advantages, we exploit the RBFs for the value function approximation. The parameters of the RBFs are usually computed based on the knowledge of the problem. However, it is possible to adapt these parameters using the observed transitions to enhance the autonomy of the method. Cross entropy and gradient descent methods were proposed by~\cite{17} for that matter. Stability of the latter was later enhanced using a restrictive technique in~\cite{18}, which is adopted in this work.

Once the structure of the value function is determined, a suitable algorithm has to be picked to train the value function approximation. Various approaches were proposed in the literature to gradually bring the value approximations close to their real values in all states. These methods were generally categorized as bootstrapping approaches (such as Fixed-Point Kalman Filter or FPKF~\cite{19}), residual approaches (e.g., Gaussian Process Temporal Difference or GPTD~\cite{20}), and projected fixed-point approaches (e.g., Least Square Temporal Difference or LSTD~\cite{21}). A comprehensive comparison of these approaches are available in~\cite{22}. Among these approaches, Kalman Temporal difference (KTD)~\cite{23} stands out as it provides not only the Minimum Mean Square Error (MMSE) estimation of the value function (given the selected structure), but also their uncertainty in terms of their error covariance which could be exploited further to reach higher sample efficiency~\cite{23}. It was also shown in~\cite{23} that GPTD is a special case of KTD. However, like other Kalman-based schemes, KTD requires the parameters of the filter (such as process and measurement noise covariances) to be known a priori, which is not the case in most practical scenarios.

Filter parameter estimation is a well-studied problem for Kalman filters and has led to numerous adaptive schemes in the literature. These methods may be roughly categorized as innovation-based adaptive methods (e.g.,~\cite{24} and multiple model adaptive schemes (e.g.,~\cite{25}). The latter has the advantage of fast adaptability when the mode of the system is changing. Most recently, the effective methods in multiple model adaptive estimation (MMAE) was discussed in~\cite{26}, where various averaging and weighting schemes were proposed and compared to achieve superior results. Multiple model approaches were previously used in RL problems. A model-based multiple model approach was introduced in~\cite{27} where the uncertainty of the system model was challenged using a set of models for the system. Moreover, a model selection approach was proposed in~\cite{28} for a multiple model KTD to overcome the filter parameter uncertainty problem, yet the models and the selection scheme were naive and therefore was not suitable for more general tasks.

In this work, various adaptive schemes are considered for value function approximation using KTD to address the parameter uncertainty problem. In particular, the paper makes the following contributions:
\begin{itemize}
\item A innovative Multiple Model Adaptive Estimation (MMAE) scheme is adopted within the RL process, referred to as the Multiple Model Kalman Temporal Differences ($\MMK$), to compensate for the lack of information about the measurement noise covariance as the most important parameter of the filter~\cite{29}, while the mean and covariance of the RBFs are updated exploiting the restricted gradient descent method of~\cite{18}.
\item To improve the sample efficiency of the proposed $\MMK$, the off-policy Q-learning is adopted to learn the optimal policy from the behaviour policy.
\item Within the proposed $\MMK$ framework, the estimated uncertainty of the value functions are exploited to form the behaviour policy leading to more visits to less certain values, therefore, improving the overall learning sample efficiency.
\end{itemize}
The proposed approach is tested over multiple RL benchmarks, which show desirable results as compared to their deep learning-based counterparts.

The remaining of the paper is organized as follows: In Section~\ref{sec:PrbFor}, the basic techniques of RL is briefly outlined. Section~\ref{sec:MM-KTD} presents the proposed $\MMK$ framework. Experimental results based on three RL benchmarks are presented in Section~\ref{sec:Sim} illustrating effectiveness and superiority of the proposed $\MMK$ framework. Finally, Section~\ref{sec:con} concludes the paper.

\section{Problem Formulation} \label{sec:PrbFor}
In this section, we briefly present the required background on RL and formulate the problem at hand. Throughout the paper, the following notations are used: Non-bold letter $X$ denotes a scalar variable, lowercase bold letter $\x$ represents a vector, and capital bold letter $\X$ denotes a matrix. Transpose of a matrix $\X$  is denoted $\X^{T}$.

\subsection{Reinforcement Learning (RL)}
In a typical RL scenario, one deals with an agent being placed in an unknown environment. At each time, the agent is considered to be in a specific state  within the available set of states denoted by $\mS$, and can take an action from the action set $\mA$ using specific policy that takes the agent to another state.
More specifically, at time step $k$, the agent at state $\s_k \in \mS$ takes an action $a_k \in \mA$ exploiting the policy $\pi_k$, which takes the system to the state $\s_{k+1} \in\mS$ with the transition probability of $P(\s_k, a_k, \s_{k+1}) \in \mP\ua$ resulting in a reward $r_k \in \mR$. The $5$-tuple $\{\mS, \mA, \mP\ua, \mR, \gamma\}$, for ($0 \leq \gamma \leq 1$) denoting the discount factor, defines the process known as the Markov Decision Process (MDP), which defines the domain of work in RL. The agent starts at an initial state denoted by $\s_0$ and continues to explore the states until it reaches a terminal state denoted by $\s_T$, where an episode is completed.

Generally speaking, the RL goal is to find a policy through a number of experimental episodes that maximizes the expected sum of discounted rewards over the future states. In achieving so, it is useful to define the state value function, $V_{\pi}(\s)$ as
\begin{eqnarray}
V_{\pi}(\s) = \mathbb{E} \left\{\sum_{k=0}^{T}\gamma^k r\k |\s_0=\s, a\k = \pi(\s\k) \right\},\label{eq:V}
\end{eqnarray}
where $\mathbb{E}\{\cdot\}$ represents the expectation function. If the value function (and also the transition probability) is known for a policy, the policy may be improved by selecting a greedy action at each step leading to the next state, which has the highest value (policy improvement)~\cite{30}. If the transition probability is unknown, it is more useful to exploit the state-action value function defined as follows
\begin{equation}
Q_{\pi}(\s, a) = \mathbb{E} \left\{\sum_{k=0}^{T}\gamma^k r\k |\s_0=\s, a_0 = a, a\k = \pi(\s\k) \right\}.\label{eq:Q}
\end{equation}
Using the state-action value function ($Q_{\pi}(\s, a)$) defined in Eq.~\eqref{eq:Q} has the advantage of direct selection of the greedy action as compared to the state value function ($V_{\pi}(\s)$) defined in Eq.~\eqref{eq:V}, where the leading states have to be identified.

\subsection{Off-Policy TD Learning}
Following the Bellman update idea~\cite{31}, sample transitions may be exploited to gradually update the value functions. In other words, at each transition (from one state to the next by taking an action), a one-step update may be performed, which is best known as a Temporal Difference (TD) update~\cite{31}. If the current policy is used to select actions for such an update, the procedure is called ``on-policy learning''. For instance, in SARSA method~\cite{32,Xia}, which is an on-policy learning method, the state-action value function is updated as follows
\begin{eqnarray}
\lefteqn{Q_{\pi}(\s\k, a\k) = Q_{\pi}(\s\k, a\k) \nonumber}\\
&&+\alpha \Big(r\k +\gamma\,Q_{\pi}(\s\nk, a\nk)  - Q_{\pi}(\s\k, a\k) \Big),
\end{eqnarray}
where $\alpha$ is the learning rate. On-policy samples are usually not very sample efficient, as the value function is learned for the current policy and not based on the optimal one. Besides, exploring new states is challenging in the on-policy methods as they follow a particular policy. On the other hand, the ``off-policy learning'' methods, also referred to as behavior policies, allow for updating the optimal policy using the information gained from other policies. In most cases, a stochastic (e.g., random) policy is selected as the behaviour policy to ensure enough exploration of new states. One of the most practiced off-policy methods is known as Q-learning~\cite{33,Ge,Xia,Li}, which updates the value function using the Bellman optimality equation as follows
\begin{eqnarray}
\lefteqn{Q_{\pi^*}(\s\k, a\k) = Q_{\pi^*}(\s\k, a\k) }\label{Eq:4}\\
&&+\alpha \Big(r\k +\gamma \max_{a\in\mA}Q_{\pi^*}(\s\nk, a)  - Q_{\pi^*}(\s\k, a\k) \Big),\nonumber
\end{eqnarray}
where $\pi^*$ represents the optimal policy. For the greedy policy, the state value function is related to the state action
value function as follows
\begin{eqnarray}
V_{\pi^*}(\s) = \max_{a\in\mA}Q_{\pi^*}(\s\k, a).
\end{eqnarray}
It is noteworthy to mention that action $a\k$ in Eq.~\eqref{Eq:4} is selected based on the behaviour policy. Once the system has converged, the optimal policy may be used as follows
\begin{eqnarray}
a\k = \arg\max_{a\in\mA}Q_{\pi^*}(\s\k, a).\label{Eq:6}
\end{eqnarray}
This completes description of off-policy TD learning. Next, we focus on value function approximation.

\subsection{Value Function Approximation}
When the number of states are finite and small, it is relatively easy to update the value function by visiting every state of the system (e.g., using Q-Learning). When the states are continuous, however, it is not viable to visit all the states, therefore, requiring to approximate the value function. Deep learning techniques provide powerful supervised learning solutions for such purposes by approximating highly nonlinear functions using labelled data. However, neural networks (used in deep learning) are notorious for problems such as over-fitting and brittleness to parameter tuning, therefore, should be used with extra care.

Alternatively, the value function may be approximated using basis functions. In this approach, the value function is estimated with a weighted sum of local basis functions, each of which is active in a local region of the state-space. The value function is then formed as follows
\begin{eqnarray}
Q_{\pi}(\s\k, a\k) = \bm{\phi}(\s\k, a\k)^T \bt\k, \label{Eq:7}
\end{eqnarray}
where $\bm{\phi}(\s, a)$ is a vector of basis functions (will be described later in Section~\ref{Sec:RBFs}) and $\bt$ is a weight vector. Various basis function may be used for such approximations, among which RBFs are proven~\cite{15} to be one of the most suitable options and are therefore selected in this work for value function approximation. This completes our background discussion on RL. Next, we present the proposed $\MMK$ framework.

\section{$\MMK$: Multiple Model Kalman Temporal Difference} \label{sec:MM-KTD}
For RL tasks with continuous state-space, the sample transitions of the system and the gained rewards are used as the data for the purpose of value function approximation and to estimate the weights. To avoid overfitting problems, supervised learning methods such as deep learning require this data to be stored and then used in batches (batch learning) for training of a system (e.g., neural networks).
Generally speaking, neural networks have considerably high memory requirements to store the input data, weight parameters, and activations as an input propagates through the network. In training, activations from a forward pass must be retained until they can be used to calculate the error gradients in the backwards pass. As an example, the $50$-layer ResNet network has $26$ million weight parameters and computes $16$ million activations in the forward pass. Measuring, roughly, the memory requirement associated with the training stage  of the ResNet-50 with a mini-batch of $32$ shows that it requires a, typically, a high performance GPU and over $7.5$ GB of local DRAM. In the contrary, sequential data processing methods such as multiple-model filters~\cite{AK1,AK2,AK3, AK4} can adapt the system with the last measurement (assuming a one-step Markov property), without the need to store the whole measurement history for learning, which results in much less memory requirement.
Using a Kalman-based approach, the posteriori of the weights $\bt\k$ can be calculated recursively using the Bayes rule as follows
\begin{eqnarray}
P(\bt\k|\Y\k) = \frac{P(\y\k|\bt\k, \Y\pk) P(\bt\k|\Y\pk)}{P(\y\k|\Y\pk)},\label{eq:Pst}
\end{eqnarray}
where $\y\k$ is the measurements vector of the system (i.e, transition information) at time step $k$ and $\Y\k$ is the set of all measurements from time Step $1$ to time Step $k$. Utilizing the probabilistic model in Eq.~\eqref{eq:Pst}, the paper proposes a Kalman-based off-policy learning scheme,  which is detailed below.

\subsection{Kalman Temporal Difference Method}
The optimal value function may be approximated from its one-step approximation using the TD method shown in Eq.~\eqref{Eq:4}, i.e.,
\begin{eqnarray}
Q_{\pi^*}(\s\k, a\k) \approx r\k +\gamma \max_{a\in \mA} Q_{\pi^*}(\s\nk, a).
\end{eqnarray}
With a change in the order of the variables, the reward at time Step $k$ may be considered as a noisy measurement from the system as follows
\begin{eqnarray}
r\k = Q_{\pi^*}(\s\k, a\k) -\gamma \max_{a\in \mA} Q_{\pi^*}(\s\nk, a) + v\k, \label{Eq:10}
\end{eqnarray}
where $v\k$ is assumed to be a zero-mean Gaussian noise with variance of $R$. In this paper, the value function is approximated as discussed in Eq.~\eqref{Eq:7}, rendering Eq.~\eqref{Eq:10} to have the
following form
\begin{eqnarray}
r\k &\!\!=\!\!& \bm{\phi}(\s\k, a\k)^T \bt\k \!-\! \gamma \max_{a \in\mA}\bm{\phi}(\s\nk, a)^T \bt\k +v\k \nonumber \\
&\!\!=\!\!& \Big[\bm{\phi}(\s\k, a\k)^T\!-\! \gamma \max_{a \in\mA}\bm{\phi}(\s\nk, a)^T\Big]\bt\k+v\k.\label{Eq:11}
\end{eqnarray}
Considering,
\begin{eqnarray}
\h\k = \bm{\phi}(\s\k, a\k) - \gamma \max_{a \in \mA} \bm{\phi}(\s\nk, a), \label{Eq:12n}
\end{eqnarray}
where $\max\limits_{a \in\mA} \bm{\phi}(\s\nk, a\k)$ is found from
\begin{eqnarray}
 \max\limits_{a \in \mA} \bm{\phi}(\s\nk, a\k)^T\bt\k,\nonumber
\end{eqnarray}
Eq.~\eqref{Eq:11} defines the measurement of the system (the reward) as a linear function of the weight function (i.e., $\bt$), which is to be estimated. Assuming the weight vector to be modelled by a linear dynamic system, i.e.,
\begin{eqnarray}
\bt\nk = \F\bt\k + \w\k, \label{Eq:13n}
\end{eqnarray}
where $\F$ is the transition matrix and $\w\k$ is a zero-mean Gaussian noise with the covariance of $\Q$, the Kalman filtering formulation may be employed to estimate the weights in a minimum MSE sense. To be more precise, the weights and their error covariance are first initialized, i.e.,
\begin{eqnarray}
\hat{\bt}_0 &=& \bt(0)\\
\text{and } \P_{\bt,0} &=& \P_{\bt}(0).
\end{eqnarray}
Then at each time step, first the weights and their error covariance are predicted as
\begin{eqnarray}
\hat{\bt}\kpk &=& \F\bt\k\\
\P_{\bt,k|k-1} &=& \F\P_{\bt,k}\F^T,
\end{eqnarray}
and then the estimations are updated using the observed reward from the transition from State $\s\k$ to the next state ($\s\nk$) as follows

\begin{eqnarray}
\K\k &\!\!=\!\!& \P_{\bt,k|k-1} \h\k\big(\h\k^T\P_{\bt, k|k-1}\h\k +R \big)^{-1}\label{Eq:18}\\
 \hat{\bt}\k &\!\!=\!\!& \hat{\bt}\kpk + \K\k\big(r\k - \h^T\k\hat{\bt}\kpk \big)\label{Eq:19}\\
\P_{\bt,k} &\!\!=\!\!& \big(\I - \K\k\h^T\k\big)\P_{\bt, k|k-1}\big(\I - \K\k\h^T\k\big)^T \label{Eq:20} \\&&
+\K\k R\K^T\k.\nonumber
\end{eqnarray}
The value function for each set of state and action is easily reconstructed through Eq.~\eqref{Eq:7}, and the optimal policy would be to select the action with the highest state-action value function at each state similar to Eq.~\eqref{Eq:6}.

When the parameters of the filter and system initial values (i.e., $\bt_0$, $\P_{\bt,0}$, $\F$, $\h\k$, $\Q$, and $R$) are known a priori, the system will provide accurate estimations. However, these values are usually not available and need to be approximated using the measurements obtained from the environment. Among these, the measurement mapping function ($\h\k$) and the measurement noise variance ($R$) are of most importance since they regulate the flow of information from new measurements. The adaptation of these parameters is the topic of the following two subsections. Other filter parameters may be selected as design parameters.

As a final note,  we would like to point out that the basic component of the constructed state-space model is the measurement equation (Eq.~\eqref{Eq:11}), which relates the reward to the basis functions. In this paper, following the formulation in Eq.~\eqref{Eq:7} we are dealing with a linear combination of the basis functions with unknown (to be estimated) weight vectors $\bm{\theta}\k$. This formulation results in a linear measurement model of Eq.~\eqref{Eq:11}. An  interesting direction for future work is to consider nonlinear measurement model~\cite{23} for estimating the value function. With regards to the dynamics of the weight vectors, at one hand, it is a common practice~\cite{AK1} to use the linear model of Eq.~\eqref{Eq:13n} when an auxiliary and unknown dynamic is introduced for the variable of interest to be estimated. Intuitively speaking, the dynamical model of Eq.~\eqref{Eq:13n} is introduced to make it possible to use state-space based solutions such as Kalman-based filters  or Particle filters.  As its true nature is unknown a-priori, the intuition is to consider it to be constant (hence having an identity type sate model $\F$) with changes being controlled by the covariance of the state model noise ($\w\k$). An interesting direction for further investigations is to learn dynamics of the introduced state model as well, e.g., using a separate neural-based module, which is the focus of our ongoing research.
\subsection{Multiple Model Adaptive Estimation}
Kalman filter is a powerful tool for accurate estimation of hidden variables when the estimation model is fully known. However, usually full knowledge about the filter parameters and initial values is not available in practical scenarios, which lead to deterioration of the system performance. Adaptive estimation is a one powerful way to remedy such a problem. In this work, a multiple model adaptation scheme~\cite{AK1, AK2, AK3, AK4} is adopted due to simplicity and effectiveness of multiple-model solutions. In such schemes, multiple candidates are considered for each of the uncertain parameters and values and the estimations made based on each set of candidates is weighted exploiting the measurement likelihood function. The number of candidates increases exponentially with the number of uncertain parameters (curse of dimensionality), therefore, it is desirable to limit the adaptation to some of the most important parameters.
In a Kalman-based estimation framework, the measurement noise variance ($R$) is one of the most important parameters and is, therefore, considered for the adaptation in this section using the multiple model technique. After observing $\s\nk$ and $r\k$ by taking action $a\k$, the measurement model ($\h\k$) is calculated. Then, different values ($R\i$) for the measurement noise variance is considered in the proposed MM-KTD scheme and a bank of mode-matched Kalman filters are implemented for adaptation of the observation noise variance. 
Eqs.~\eqref{Eq:18}-\eqref{Eq:20} are, therefore, replaced with the following
\begin{eqnarray}
\K\i\k &\!\!=\!\!& \P_{\bt,k|k-1} \h\k\big(\h\k^T\P_{\bt, k|k-1}\h\k +R\i \big)^{-1}\label{Eq:21}\\
\hat{\bt}\i\k &\!\!=\!\!& \hat{\bt}\kpk + \K\i\k\big(r\k - \h\k^T \hat{\bt}\kpk\big)\\
\P\i_{\bt,k} &\!\!=\!\!& \big(\I - \K\i\k\h\k^T\big)\P^T_{\bt, k|k-1}\big(\I - \K\i\k\h\k^T\big) \nonumber\\ \lefteqn{+\K\i\k R\i\K^{(i)^T}\k,\label{Eq:23}}
\end{eqnarray}
where superscription $i$ denotes the Kalman filter value for the $i^{\text{th}}$ filter which exploits $R\i$ as its measurement noise covariance.
The posteriori of each mode-matched filter is weighted based on its associated and normalized  weight, which are then combined to form the system's posteriori, i.e.,
\begin{eqnarray}
P(\bt\k|\Y\k) =\sum_{i=1}^{M}\omega\i P(\bt\k|\Y\k, R\i), \label{Eq:24}
\end{eqnarray}
where $M$ is number of mode-matched filter within the $\MMK$ framework and,
\begin{eqnarray} \label{Eq:25}
\lefteqn{\omega\i = P(r\k|\bt\kpk,R\i)=}\\
&& \!\! \!\!\!\!\!\!\!\!\! \!\!\!\!\!\!\!\!\!  c.e^{\frac{-1}{2}\big(r\k-\h\k^T\hat{\bt}\kpk\big)^T\big(\h\k^T\P_{\bt,k|k-1}\h\k+R\i\big)^{-1}\big(r\k-\h\k^T\hat{\bt}\kpk\big)\nonumber},
\end{eqnarray}
where $w\i$ is the normalized measurement likelihood for the $i^{\text{th}}$ filter and,
\begin{eqnarray} \label{Eq:26}
c &=& \frac{1}{\sum_{i=1}^{M} w\i},
\end{eqnarray}
%
Using Eq.~\eqref{Eq:24}, the weight and its error covariance are then updated as follows
\begin{eqnarray}
\!\!\!\!\!\!\!\!\!\!\!\!\hat{\bt}\k &\!\!=\!\!& \sum_{i=1}^{M}\omega\i\hat{\bt}\i\k\\
\!\!\!\!\!\!\!\!\!\!\!\!\P_{\bt,k} &\!\!=\!\!& \sum_{i=1}^{M}\omega\i\left(\P_{\bt,\k}\i + (\hat{\bt}\i-\hat{\bt})(\hat{\bt}\i-\hat{\bt})^T \right).\label{Eq:New28}
\end{eqnarray}
%
As a final note, it is interesting to highlight the connections between the above multiple model frame work of the proposed MM-KTD with other methods, in particular gating approach of Mixture of Experts (MoE)~\cite{Hinton:2017}. In a general setting, the goal of the weight update step in a multiple-model framework is to find the
conditional probability density function (PDF) corresponding to mode $i$, denoted by $m^{i}$ for ($1 \leq i \leq M$), given all the observations up to the current time,~i.e.,
\begin{eqnarray}\label{eq:modelProb3}
\!\!\!\!\!\!\!\!p\big(m^{(i)}\k|\Y\k\big) &\!\!\!\!\!=\!\!\!\!&  \frac{p\big(\y\k| \Y\pk, m^{(i)}\k\big) p\big(m^{(i)}\k|\Y\pk\big) }
{\sum_{j=1}^{M} p\big(\y\k| \Y\pk, m\k^{(j)}\big) p\big(m\k^{(j)}|\Y\pk\big)},
\end{eqnarray}
where the denominator  is a normalizing factor to ensure that $p(m^{(l)}\k|\Y\k)$ is a proper  PDF. Term  $\aL\i\k \triangleq p(\y\k|\Y\pk, m^{(i)}\k)$ in  the nominator of Eq.~\eqref{eq:modelProb3} is the  likelihood function of mode $i$, which is proportional to the exponential term in Eq.~\eqref{Eq:25}. Therefore, the corresponding weight for the filter matched to mode $i$, for ($1 \leq i \leq M$), can be simplified~as
\begin{eqnarray}\label{eq:core2}
\omega\i\k  \triangleq p(m\k^{(i)}|\Y\k) = \frac{\omega\i\pk\aL\i\k}{\sum_{j=1}^{M}\omega^{(j)}\pk\aL^{(j)}\k}.
\end{eqnarray}
On the contrary, the MoE, typically, uses the soft max adaptation for Gaussian gating to obtain the associated weight, denoted here by $\omega^{(i, MoE)}\k$, for each model (expert). The weight is obtained by multiplying the input (in our case, the estimated state vector $\hat{\bt}\k\i$ by model $i$, for ($1 \leq i \leq M$)) by a trainable weight matrix $\W_g$ and then applying the Softmax function as follows
\begin{eqnarray}
\omega^{(i, MoE)}\k = \frac {\exp\big(\hat{\bt}\i\k . \W_g\big)}{\sum _{j=1}^{M}\exp\big(\hat{\bt}^{(j)}\k . \W_g\big)}.\label{eq:New31}
\end{eqnarray}
Comparing Eq.~\eqref{eq:New31} with Eq.~\eqref{eq:core2} reveals the potentials of the multiple model approach in comparison to the MoE. This completes our discussion on multiple model adaptive estimation. Next, we discuss the update process for computation of the basis functions.
\subsection{Basis Function Update}\label{Sec:RBFs}
Vector $\h\k$, defined in Eq.~\eqref{Eq:12n}, is the measurement mapping function and is required to be computed for accurate estimations within the context of Kalman-based filtering schemes. Such a prior knowledge, however, is typically not available, therefore $\h\k$ should be adapted to its correct value. Since $\h\k$ is formed by the basis functions, its adaptation necessitates the adaptation of the basis functions. The vector of basis functions shown in Eq.~\eqref{Eq:7} is formed as follows,
\begin{eqnarray}
\bm{\phi}(\s\k,a\k) = [\phi_{1,a_1},\ldots \phi_{N,a_1}, \phi_{1,a_2}, \ldots \phi_{N,a_D}]^T,\label{Eq:28}
\end{eqnarray}
where $N$ is the number of basis functions per action and $D$ is the total number of actions (the arguments of the basis functions are omitted for brevity). Each basis function ${\phi}_{n,a_d}(\s\k,a\k)$ is selected as a RBF, which is defined based on its mean vector $\bmu_{n,a_d}$ and covariance matrix $\Sig_{n,a_d}$ as follows
\begin{eqnarray}
\bm{\phi}_{n, a_d}(\s\k,a\k) = e^{\frac{-1}{2}(\s\k-\u_{n, a_d})\bm{\Sigma}^{-1}_{n,a_d}(\s\k-\u_{n, a_d})},\label{Eq:29}
\end{eqnarray}
where $\u_{n, a_d}$ and $\bm{\Sigma}_{n, a_d}$ are the mean and covariance of this radial basis function. Due to the large number of parameters associated with the measurement mapping function (i.e., $2 \times N \times D$), it is reasonable to adapt these parameters through a gradient descent-based adaptation scheme rather than the multiple model method. For this purpose, the Restricted Gradient Descent (RGD) method proposed in~\cite{18} is adopted in this work. Using RGD, first the gradient of the object function with respect to the parameters of each basis function is calculated using partial derivations. The goal is to minimize the objective function, which is defined as the difference between the estimated value function and its one-step (TD) update, i.e.,
\begin{eqnarray}
\S\k = \left( Q_{\pi^*}(\s\k, a\k) -r\k -\gamma\max_{a \in \mA} Q_{\pi^*}(\s\nk, a)\right)^2. \label{Eq:30}
\end{eqnarray}
The gradient of the object function with respect to the parameters of the RBFs is calculated using the chain rule,
\begin{eqnarray}
\Delta\u &=& -\frac{\partial\S\k}{\partial\u} =  -\frac{\partial\S\k}{\partial Q_{\pi^*}}\frac{\partial Q_{\pi^*}}{\partial \bm{\phi}}\frac{\partial \bm{\phi}}{\partial\u}\\
\text{and }\Delta\Sig &=& -\frac{\partial\Sig\k}{\partial\u} =  -\frac{\partial\S\k}{\partial Q_{\pi^*}}\frac{\partial Q_{\pi^*}}{\partial\bm{\phi}}\frac{\partial\bm{\phi}}{\partial\Sig},
\end{eqnarray}
where the partial derivations are calculated using Eqs.~\eqref{Eq:7},~\eqref{Eq:29}, and \eqref{Eq:30} as follows
\begin{eqnarray}
\frac{\partial\S\k}{\partial Q_{\pi^*}} &=& 2\S^{\frac{1}{2}} \label{Eq:34}\\
\frac{\partial Q_{\pi^*}}{\partial\bm{\phi}} &=& \bt^T\k\\
\frac{\partial\bm{\phi}}{\partial\u} &=& \bm{\phi}\Sig^{-1} (\s\k -\u)\\
\frac{\partial\bm{\phi}}{\partial\Sig} &=& \bm{\phi}\Sig^{-1} (\s\k -\u) (\s\k -\u)^T\Sig^{-1}.
\end{eqnarray}
The mean and covariance of the RBFs are then adapted using the calculated partial derivative as follows
%
\begin{eqnarray}
\lefteqn{\u_{n,a_d} =  \u_{n,a_d} +\lambda_{\u}\Delta\u\nonumber}\\
&=\!\!\!& \u_{n,a_d} -2\lambda_{\u}\S^{\frac{1}{2}} Q_{\pi^*}\Sig^{-1}(\s\k- \u_{n,a_d})\label{Eq:37}\\
\lefteqn{\Sig_{n,a_d} =  \Sig_{n,a_d} +\lambda_{\Sig}\Delta\Sig\nonumber}\\
&=\!\!\!& \Sig_{n,a_d} -2\lambda_{\Sig}\S^{\frac{1}{2}} Q_{\pi^*}\Sig^{-1}_{n, a_d}(\s\k- \u)(\s\k- \u)^T\Sig_{n,a_d}^{-1},\nonumber\\\label{Eq:38}
\end{eqnarray}
where $\lambda_{\u}$ and $\lambda_{\Sig}$ are the adaptation rates. To make the system more stable, only one of the updates shown in Eqs.~\eqref{Eq:37} and~\eqref{Eq:38} will be performed as discussed in~\cite{18}. To be more precise, when the size of the covariance is decreasing (i.e., $\S^{\frac{1}{2}}Q_{\pi^*} > 0$), the covariances of the RBFs are updated using
Eq.~\eqref{Eq:38}, otherwise, their means are updated using Eqs.~\eqref{Eq:37}. Using this approach, unlimited expansion of the RBF covariances is avoided.

\subsection{Active Learning}
In order to ensure enough exploration of the  states in off-policy learning methods, the behaviour policy is usually chosen to be a stochastic policy (e.g., a random policy). However, such a choice commonly leads to sample inefficiency, which is already a big problem in model-free RL methods. One advantage that the proposed $\MMK$ learning framework offers over other optimization-based techniques (e.g., gradient descent-based methods) is the calculation of the uncertainty for the weights ($\P_{\bt}$), which is directly related to the uncertainty of the value function. This information can then be used at each step to select the actions, which  lead to most reduction in the uncertainty of the weights. Using the information of the Kalman filter (information filter~\cite{AK2}), the information of the weights, which is denoted by the inverse of $\P_{\bt}$ is updated as follows
\begin{eqnarray}
\P^{-1}_{\bt,k} = \P^{-1}_{\bt,\kpk} +\h\k R^{-1}\h\k^T. \label{Eq:39}
\end{eqnarray}
Since only the second element (i.e., $\h\k R^{-1}\h^T\k$) in the right hand side of Eq.~\eqref{Eq:39} is affected by the choice of the action (as it changes $\h\k$), the action is selected such that this term is maximized. More specifically, the action is obtained by maximizing the information of the weights, i.e.,
\begin{eqnarray}
a\k &=& \arg\max_a \Big(\h\k (\s\k,a)R^{-1}\h^T\k(\s\k, a) \Big)\nonumber\\
&=&\arg\max_a \Big(\h\k (\s\k,a)\h^T\k(\s\k, a) \Big).\label{Eq:40}
\end{eqnarray}
The second equality in Eq.~\eqref{Eq:40} is constructed as $R$ is a scalar. The proposed behavior policy in Eq.~\eqref{Eq:40} is different from that of Reference~\cite{23}, where a random policy was introduced, which favored actions with less certainty of the value function. Even though favoring the actions that reduce the uncertainty of the value function is a good idea, the random nature of such policies make them less sample efficient than expected. The proposed $\MMK$ framework is briefed in Algorithm~\ref{algo:1}.
It is worth further clarifying computation of Step 6 in Algorithm~\ref{algo:1}. For learning the optimal policy, the control action $a_k$ is selected based on the behavioral policy, which leads to most reduction in the uncertainty of the weights in Eq.~\eqref{Eq:40}. Once the system has been converged, the resulted optimal policy will be used based on Eq.~\eqref{Eq:6} to  select the actions during the testing phase. Because the state-space is continuous, we have approximated the value function of Eq.~\eqref{Eq:6} using RBFs. The value function is estimated with a weighted sum of RBFs $\bm{\phi}(\s\k,a\k)$ with the weight vectors $\bm{\theta}\k$. For estimating the weights, the sample transition of the system, i.e., $\s\k, a\k$ and the gained reward $r\k$ are used in a Kalman-based approach as in Eqs.~\eqref{Eq:11}, and~\eqref{Eq:13n}. Therefore, the control action in Step 6 will be found based on the proposed active learning behavioral policy in Eqs.~\eqref{Eq:12n}, and~\eqref{Eq:40}. Because the matrix $\big(\h\k (\s\k,a)\h\k^T(\s\k, a)\big)$ that generates the control action cannot be maximized, we have maximized its trace $\big(\h\k^T (\s\k,a)\h\k(\s\k, a)\big)$. Finally, we note that the proposed MM-KTD algorithm is designed for systems with finite number of actions. It is worth mentioning that having infinite number of actions per state is typical of continuous control tasks~\cite{Li, Al-Tamimi:2007}. Extension of the proposed framework for application to infinite-dimensional action space is an interesting direction for future research work.
\begin{algorithm}[!t]
\caption{\textproc{The Proposed $\MMK$ Framework}}
\label{algo:1}
\begin{algorithmic}[1]
\State \textbf{Learning Phase:}
\State Set $\bt_0, \P_{\bt,0},\F,\u_{n,i_d}, \Sig_{n,i_d}$ for $n={1,2,\hdots,N}$ and $i_d={1,2,\hdots,D}$
\State  \textbf{Repeat} (for each episode):
\State \quad Initialize $\s\k$
\State \quad \textbf{While} $\s\k \neq \s_T$ \textbf{do}:
\State \quad \quad $a\k=\arg\max\limits_a \Big(\h\k (\s\k,a)\h\k^T(\s\k, a) \Big)$
\State \quad \quad Take action $a\k$ , observe $\s_{k+1}, r\k$
\State \quad \quad Calculate $\bm{\phi}(\s,a)$ via Eqs.~\eqref{Eq:28} and~\eqref{Eq:29}
\State \quad \quad $\h\k(\s\k,a\k)\!=\!\bm{\phi}(\s\k,a\k)-\gamma\arg\max\limits_a \bm{\phi}(\s_{k+1},a)$
\State \quad \quad  $\hat{\bt}\kpk=\F \hat{\bt\k}$
\State \quad \quad $\P_{\bt,k|k-1}=\F \P_{\bt,k-1} \F^T +\Q$
\State \quad \quad \textbf{for} $i=1:M$ \textbf{do}:
\State \quad \quad \quad $\bm{k}\i\k=\P_{\bt,k|k-1} \h\k (\h^T\k \P_{\bt,k|k-1}\h\k+R\i)^{-1}$
\State \quad \quad \quad $\hat{\bt}\k\i=\hat{\bt}\kpk+\bm{k}\k\i (r\k-\h^T\k \hat{\bt}\kpk)$
\State  \quad \quad \multiline{$\P_{\bt,k}\i\!\!\!\!=\!\!\!\!(\I\!-\!\bm{K}\i\k \h^T\k)\P_{\bt,k|k-1} (\I-\bm{K}\i\k \h^T\k)^T+\bm{K}\i\k R\i {\bm{K}\i\k}^T $}
\State \quad \quad \textbf{end for}
\State \quad \quad \multiline{ Compute $c$ and $w\i$ by  using $\sum_{i=1}^{M} w\i=1$ \&~\eqref{Eq:25}}
\State \quad \quad $\hat{\bt}\k=\sum_{i=1}^{M} w\i \hat{\bt}\k\i $
\State \quad \quad $\P_{\bt,\k} = \sum_{i=1}^{M}\omega\i\left(\P_{\bt,\k}\i + (\hat{\bt}\i-\hat{\bt})(\hat{\bt}\i-\hat{\bt})^T \right)$
\State \quad \quad \textbf{RBFs Parameters Update:}
\State \quad \quad $S\k = \left( Q_{\pi^*}(\s\k, a\k) -r\k -\gamma\max\limits_{a \in \mA} Q_{\pi^*}(\s\nk, a)\right)^2$
\State \quad \quad \textbf{if} $S^{\frac{1}{2}}Q_{\pi^*} > 0$ \textbf{then}:
\State \quad \quad \quad Update $\Sig_{n,a_d}$ via Eq.~\eqref{Eq:37}
\State \quad \quad \textbf{else:}
\State \quad \quad \quad Update $\u_{n,a_d}$ via Eq.~\eqref{Eq:38}
\State \quad \quad \textbf{end if}
\State \quad \textbf{end while}
\State \textbf{Testing Phase:}
\State \quad \textbf{Repeat} (for each trial episode):
\State \quad \quad \textbf{While} $\s\k \neq \s_T$ \textbf{do}:
\State \quad \quad $a\k=\arg \max\limits_{a}\bm{\phi}(\s\k, a)^T \bt\k$
\State \quad \quad Take action $a\k$, and observe $\s\nk,r\k$
\State \quad \quad \textbf{End While}
\end{algorithmic}
\end{algorithm}
\section{Experimental Results}  \label{sec:Sim}
\begin{figure}[t!]
\centering
\includegraphics[scale=0.6]{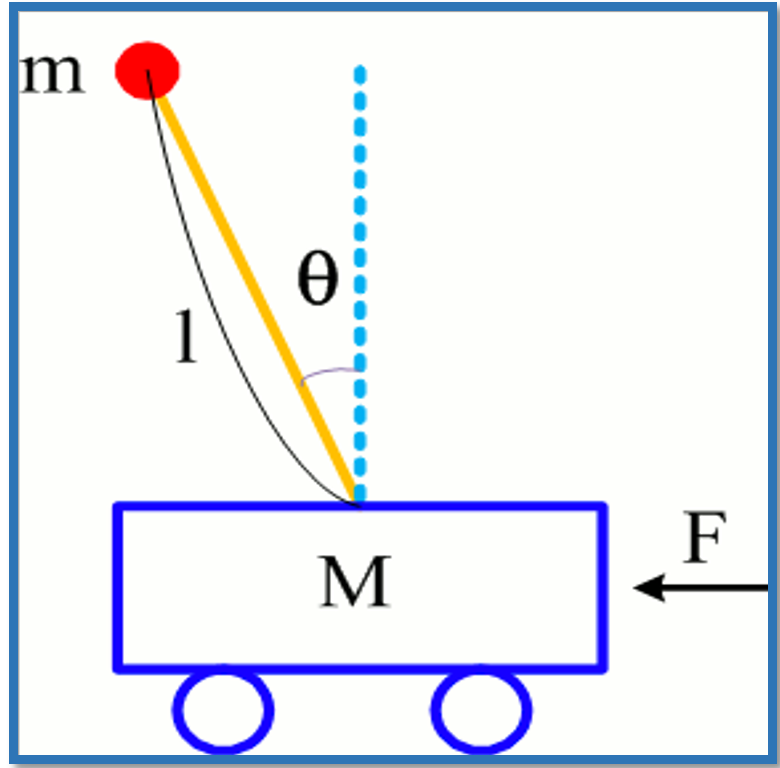}
\caption{\small The Inverted Pendulum platform.}\label{Fig:1}
\end{figure}
In this section, we evaluate performance of the proposed $\MMK$ framework.
In order to demonstrate the effectiveness and sample efficiency of the $\MMK$, which is a model-free and multiple model RL scheme,
three popular RL benchmarks, i.e., Inverted Pendulum, Mountain Car, and Lunar Lander are considered, experimented and different comparisons are performed.\footnote{Implementations of the MM-KTD model for all three RL benchmarks is available publicly for open access at https://github.com/parvin95/MM-KTD.} 
For the implementation of the proposed MM-KTD, a hardware with a $2.6$ GHz Intel Core i7 processor and $12$GB  RAM has been used. 
To have a fair comparison, we used optimized parameters for the NFQ method, and the KTD approach as specified in~\cite{6, 23}, respectively. It is also worth mentioning that one benefit of the proposed MM-KTD framework (which comes from its multiple-model architecture) is its superior ability to deal with scenarios where enough information about the underlying parameters is not fully available. 

\subsection{Inverted Pendulum}  \label{sec:cart_pole}
In the first experiment, the Inverted Pendulum platform is considered, which is shown in Fig.~\ref{Fig:1}. The weight of the base is $8$ kg, while the weight of the pendulum is assumed to be $2$ kg. The length of the pendulum is $0.5$ m. The pendulum is initially left at an angle (arbitrarily close to the upright position) and then its base (object with mass of $8$ kg) is moved to keep its balance. The base may be moved to the left or right with a force of
$50$N, or not moved. The episode ends once the pendulum is fallen behind a horizontal line. The goal is to prevent the pendulum from falling below the horizontal line as long as possible. The state of the system is determined as the pendulum angle from the upright position and its angular velocity (i.e., $\s\k = [\theta, \dot{\theta}]^T$). At each time step, given the state ($\s\k$) and the taken action ($a\k$), the next state of the system is determined through the dynamics of the system (which is not known to the agent). If the angle of the pendulum in the next state is beyond the horizontal line (i.e., $|\theta| >\pi/2$), then a reward of $-1$ will be fed back to the agent. Otherwise, the reward is kept as $0$.

\begin{figure}[t!]
\centering
\includegraphics[scale=0.55]{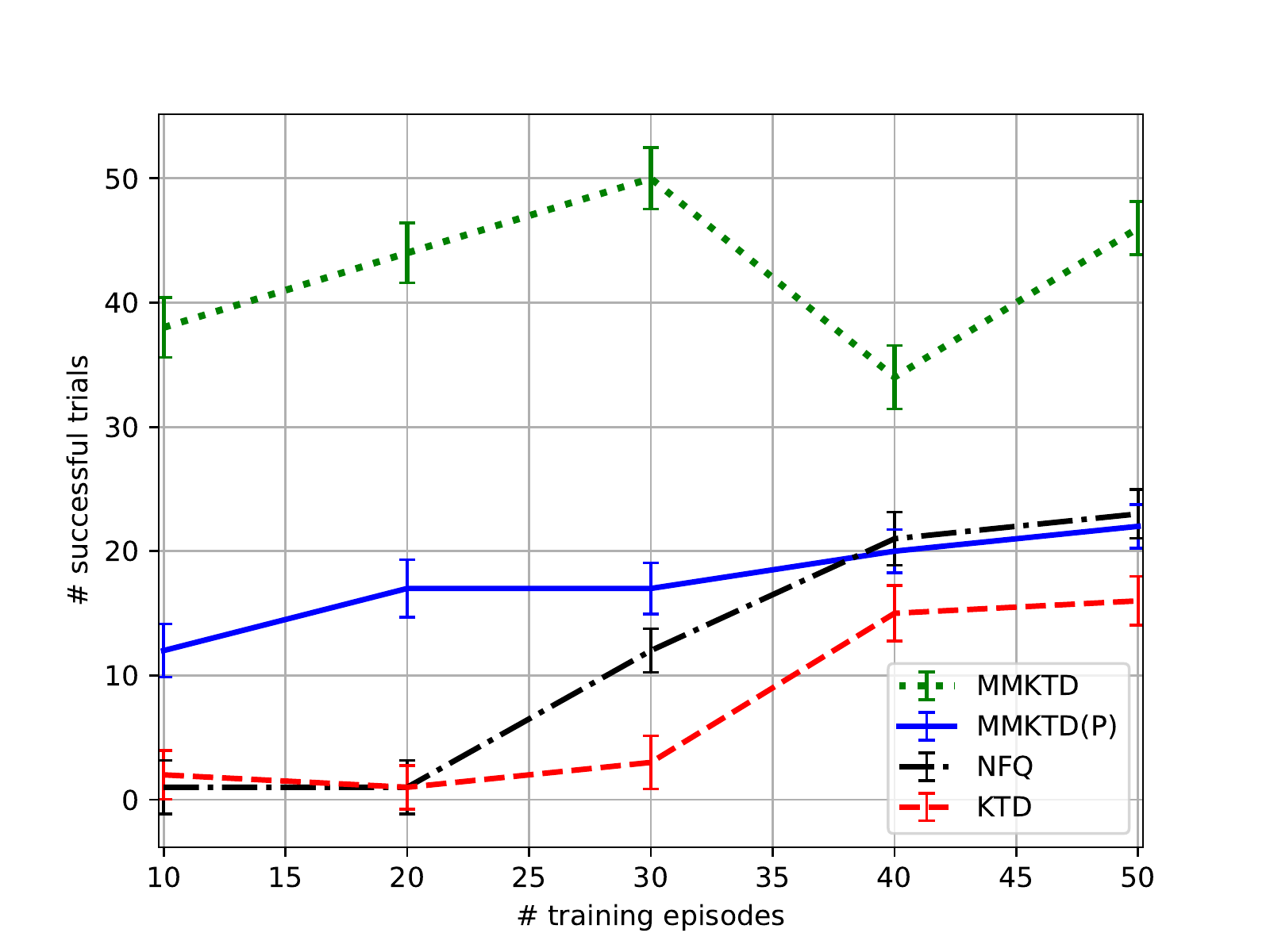}
\caption{\small The average (lines) and 95\% confidence interval (error bars) of the number of successful trials over 50 trials based on the proposed MM-KTD scheme as compared with other RL methods for the Inverted Pendulum environment.}\label{Fig:2}
\end{figure}
The proposed $\MMK$ is employed in this problem using $9$ RBFs and a bias parameter. Since there are three possible actions (i.e., $\mA =\{-50, 0, +50 \}$), the size of the feature vector is $30$. The mean and covariance of the RBFs are initialized as follows
\begin{eqnarray}
\u_{n,a_d} &\in& \{-\pi/4, 0, +\pi/4 \}\times \{-0.5, 0, +0.5\},\label{Eq:mean}\\
\Sig_{n,a_d} &=& \I_2,
\end{eqnarray}
where $\I_2$ is the identity matrix of size ($2\times 2$). The vector of basis functions in Eq.~\eqref{Eq:28} are, therefore, given by
\begin{eqnarray}
\bm{\phi}(\s\k,a\k=+50)\!=\! [1,\phi_{1,a_d},\ldots ,\phi_{9,a_d}, 0, \ldots 0,0,\ldots, 0]^T \\
\bm{\phi}(\s\k,a\k=-50)\! = \! [0, \ldots ,0,1,\phi_{1,a_d},\ldots \phi_{9,a_d},0, \ldots ,0 ]^T\\
\bm{\phi}(\s\k,a\k=0)\!=\![0, \ldots ,0,0, \ldots ,0,1,\phi_{1,a_d},\ldots \phi_{9,a_d}]^T,
\end{eqnarray}
where the value of $\phi_{n,a_d}$ for $n \in \{1,2,\dots,9\}$ is calculated via Eq.~\eqref{Eq:29}.
The initial values of $\lambda_{\u}$ and $\lambda_{\Sig}$ are selected as $200$ and $100$, respectively by using trial and error to keep the system stable. The time step is selected to be $10$ milliseconds and the discount factor is selected as $0.95$, (i.e.,  $\gamma = 0.95$). The process noise covariance is selected small ($\Q\k = 10^{-3}\I_{30}$), and the transition matrix is selected to be the identity matrix ($\F = \I_{30}$). The measurement noise covariance candidates are selected from the following set,
\begin{eqnarray}
R\i \in \{0.01,0.1, 0.2, 0.5, 1, 2, 5, 10, 20, 50, 100\}.
\end{eqnarray}
The initial weights are selected to be zero, (i.e., $\bt_0 = \bm{0}_{30}$), while the initial error covariance is chosen as $\P_{\bt,0} = 10\I_{30}$. Each experiment is started from an angle randomly selected from a normal distribution with mean zero and standard deviation of $0.1$. The agent is first trained through a set of episodes, then tested in $50$ episodes to find if it can keep the pendulum in balance for at least $5$ seconds ($500$ samples). Various number of episodes are used for training the system. To highlight the effectiveness of the proposed $\MMK$, the achieved results are compared with that of KTD method of~\cite{23}, $\MMK$ with no active learning (denoted by $\MMK$ (P)), and Neural Fitted Q (NFQ) learning method of~\cite{6}, 
which performs the update session by considering an entire set of transition experiences in an off-line fashion, instead of using an online approach for updating the neural value function. More specifically, the NFQ scheme optimizes the sequence of the loss function using Rprop algorithm, which is a supervised learning method for batch learning to update the parameters of the Q-network. The available historical observations without performing any exploration that is used in the NFQ leads to a substantially lower amount of training experience required for learning optimal policies. In this approach, we just have a loop over trained steps. The NFQ uses the current policy in order to get the target value, while the DQN uses the target network to achieve the goal. As the nature of this paper is to use the proposed algorithm to restrict the number of learning episodes with reduced number of training data, which is critical for practical application in real scenarios, we focused only on comparison with algorithms that have the least required number of learning episodes. 

Each testing scenario is repeated $10$ times for a specific number of training episodes to minimize the randomness of the achieved number of successful trials out of total $50$ trials. For evaluation purposes, we formed the average and $95$\% confidence interval of the number of successful trials out of the $10$ repetitions. Fig.~\ref{Fig:2}  shows the resulted mean (lines) and 95\% confidence interval (error bars) for different number of training episodes ranging from $10$ to $50$, which are also briefed in Table~\ref{Table:1}. 
As it was expected, the proposed $\MMK$ is the most sample efficient algorithm of all. In addition, the proposed method offers the highest asymptotic performance in the performed experiments. This superior performance comes as a result of adaptive estimation of the value function through active Q-learning. It is expected that the performance of $\MMK$ (P) and NFQ become better with more training episodes and get closer to that of the proposed $\MMK$. However, original KTD cannot reach that level of performance as it lacks accurate knowledge of the filter's parameters.
\begin{table}[t]
\vspace{-.1in}
\caption{\small The number (percentage) of successful trials of the proposed MM-KTD scheme as compared with other RL methods for the Inverted Pendulum environment.}\label{Table:1}
\centering
\begin{tabular}{|c|p{1cm}|p{1.6cm}|p{1.1cm}|p{1.1cm}|}
\hline
\multirow{2}[2]{*}{\textbf{\!\!\!\# Training Episodes\!\!\!}} & \multirow{2}[2]{*}{\textbf{KTD}} &  \multirow{2}[2]{*}{\textbf{$\MMK$ (P)}} &  \multirow{2}[2]{*}{\textbf{$\MMK$}}&  \multirow{2}[2]{*}{\textbf{NFQ}} \\
&&&&\\
\hline
\!\!\!\multirow{2}[2]{*}{\textbf{10}}\!\!\! & 2.1/50 (4.2\%) &  12.3/50 (24.6\%) &  38.5/50 (77\%) &  1.9/50 (3.8\%) \\
\hline
\!\!\!\multirow{2}[2]{*}{\textbf{20}}\!\!\! & 1/50 (2\%) &  17.4/50 (34.8\%) &  44/50 (88\%) &  1.5/50 (3\%) \\
\hline
\!\!\!\multirow{2}[2]{*}{\textbf{30}}\!\!\! & 3.6/50 (7.2\%) &  17.6/50 (35.2\%) &  50/50 (100\%) & 21.6/50 (43.2\%) \\
\hline
\!\!\!\multirow{2}[2]{*}{\textbf{40}}\!\!\! & 15/50 (30\%) &   20.7/50 (41.4\%) &  34/50 (68\%) & 15/50 (30\%)\\
\hline
\!\!\!\multirow{2}[2]{*}{\textbf{50}}\!\!\! & 16.5/50 (33\%) &   22.0/50 (44\%) &  46.6/50 (93.2\%) & 23/50 (46\%)\\
\hline
\end{tabular}
\end{table}
\begin{figure}[t!]
  \begin{subfigure}{4cm}
    \centering
    \includegraphics[width=4.5cm]{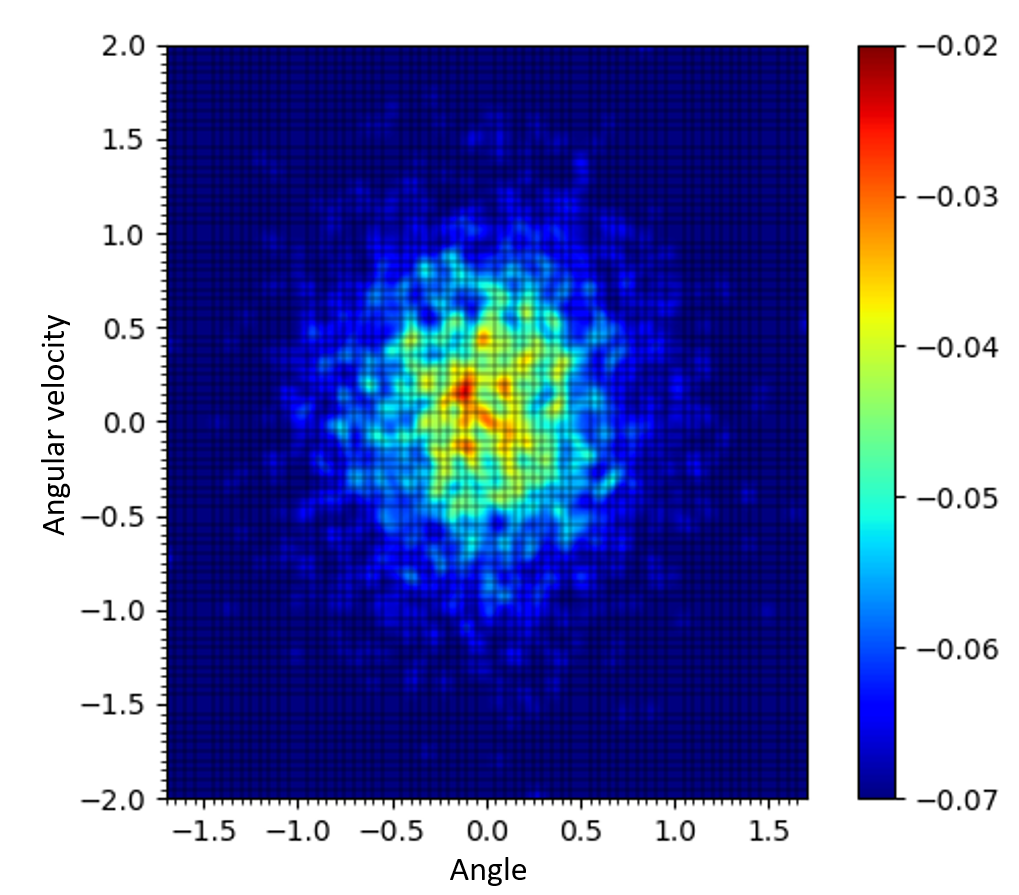}
    \caption{}
  \end{subfigure}
  \begin{subfigure}{4cm}
    \centering
    \includegraphics[width=4.5cm]{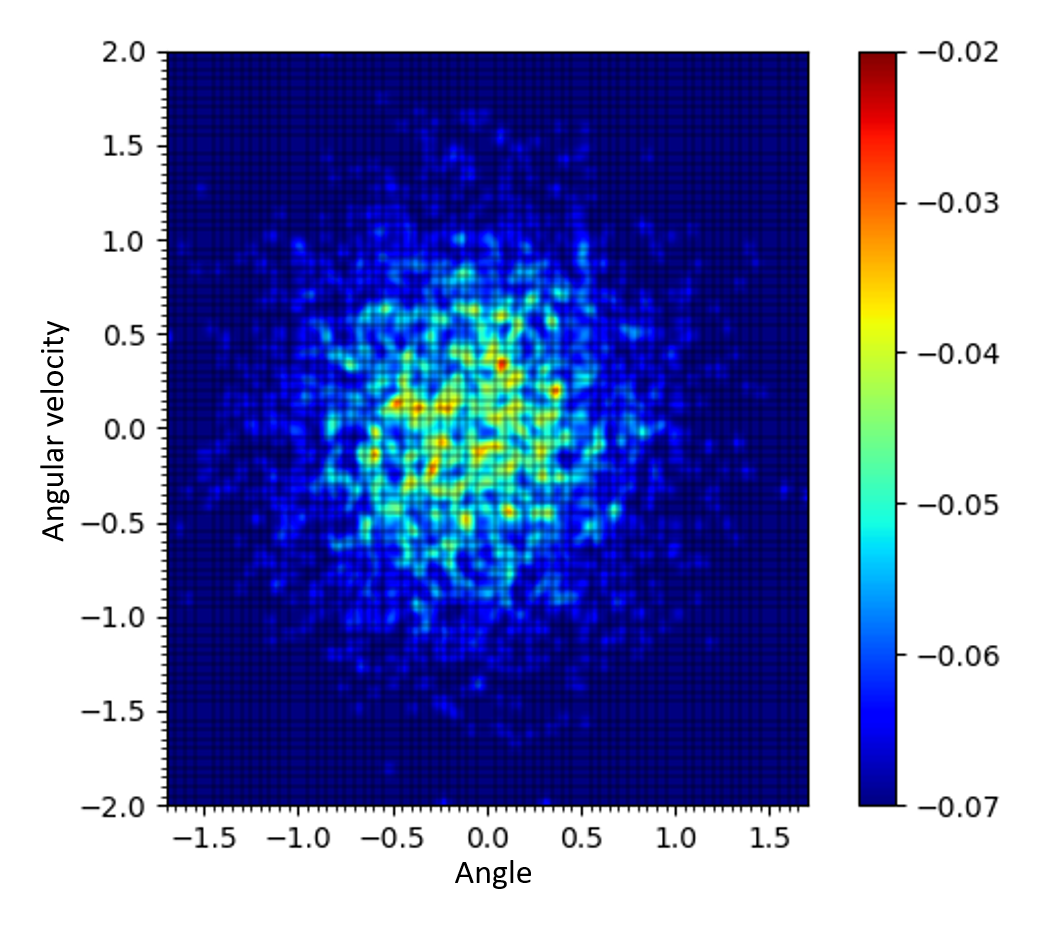}
    \caption{}
  \end{subfigure}
 \\
  \begin{subfigure}{4cm}
    \centering
    \includegraphics[width=4.5cm]{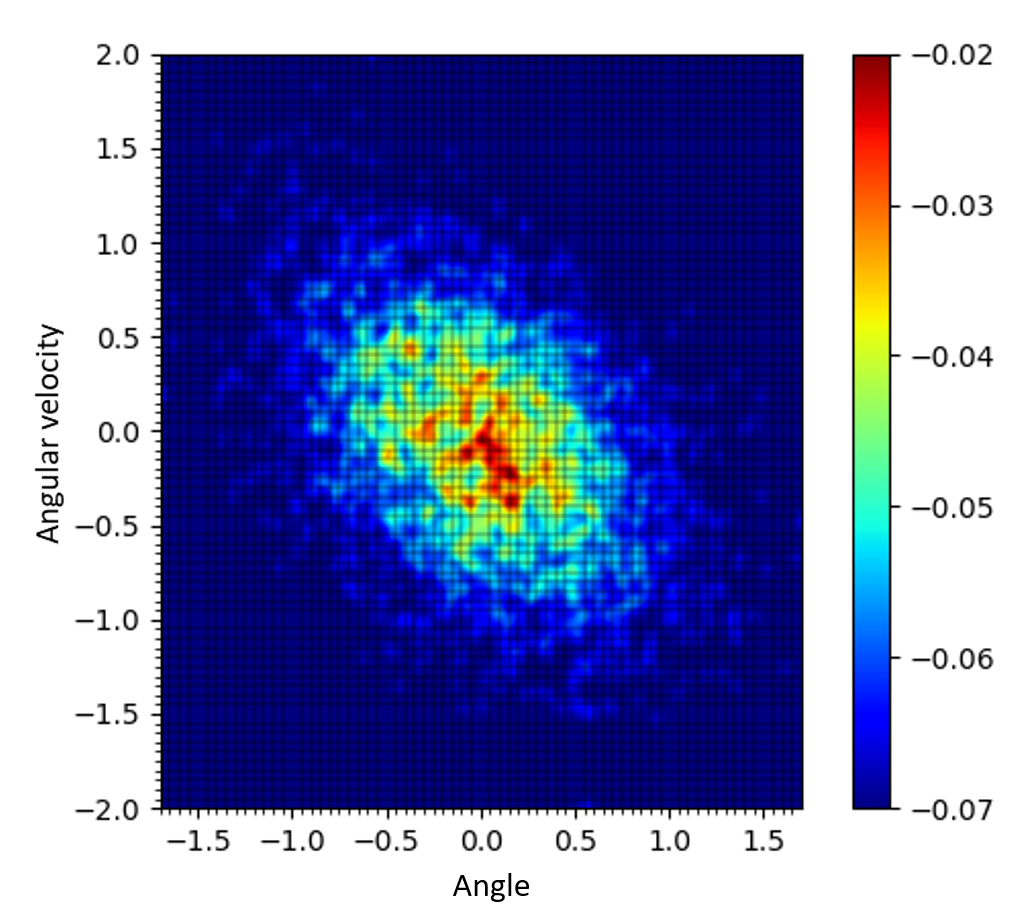}
    \caption{}
  \end{subfigure}
  \begin{subfigure}{4cm}
    \centering
    \includegraphics[width=4.5cm]{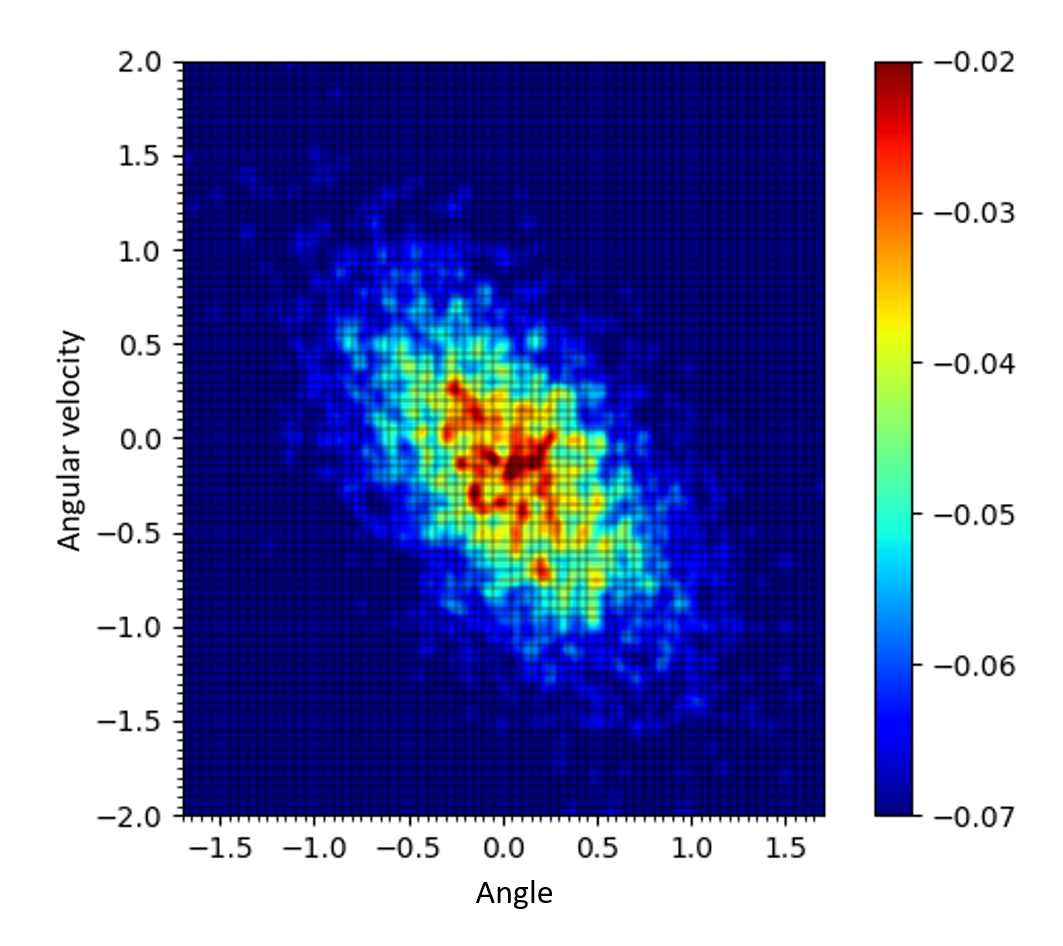}
    \caption{}
  \end{subfigure}
\caption{\small The state value function of the greedy policy ($V_{\pi^*}(\s)$) after $10$ episodes of training using: (a) $\MMK$  (b) $\MMK$ (P) (c) NFQ, and (d) KTD}\label{Fig:3}
\end{figure}

The state value function of the methods after 10 training episodes are depicted in Fig.~\ref{Fig:3}, where the $x$ and $y$ axes represent $\theta$ and $\dot{\theta}$ of the pendulum, respectively. 
It is worth mentioning that when the state value function has the higher values around the vertical (origin) position of the Inverted Pendulum (states close to $\theta=\dot{\theta}=0$), it means that the selected actions, resulted from the behavioral policy, have led to the states near the vertical position, which is the expected result, i.e., the pendulum is above the horizontal line. Therefore, based on Fig.~\ref{Fig:3}, the proposed active learning method causes higher sample efficiency of $\MMK$ scheme compared to the other techniques. 
Due to the higher sample efficiency, $\MMK$ can quickly concentrate higher values for states closer to the origin. However, other practiced methods are still far from this stage, therefore, fail to perform in an acceptable fashion with restricted amount of experience.

\begin{figure}[t!]
\centering
\includegraphics[scale=0.45]{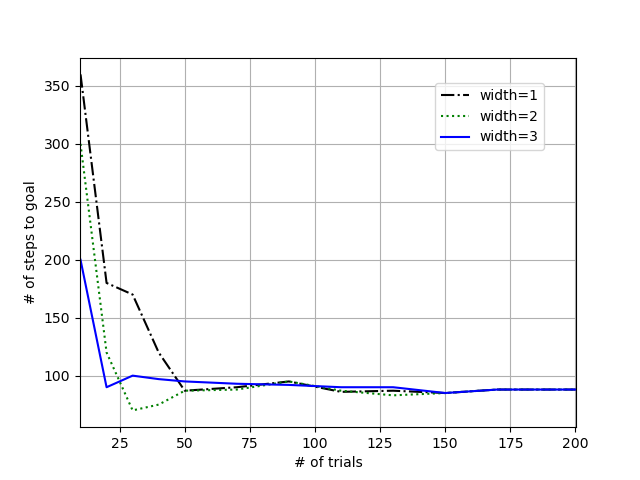}
\caption{\small Stability analysis of the RBFs.}\label{Fig:stab}
\end{figure}

Finally, to evaluate stability of the utilized RBFs, we have conducted the following Monte-Carlo (MC) study. In particular, we have fixed the number of RBFs to be $9$ and their means as selected based on Eq.~\eqref{Eq:mean}. Then, the proposed MM-KTD scheme has been performed on the Inverted Pendulum task for $200$ different trials. The entire process has been repeated $100$ times (i.e., through a MC simulation of $100$ runs) using three different values of the widths $(\Sig)$ of the RBFs. The number of steps to the goal was averaged over the $100$ runs. Fig.~\ref{Fig:stab} illustrates potential stability of the utilized RBFs. As can be observed from Fig.~\ref{Fig:stab},  steady state performance can be achieved  by using the RBFs for the value function approximation.

\subsection{Mountain Car}  \label{sec:car}
\begin{figure}[t!]
\centering
\includegraphics[scale=0.55]{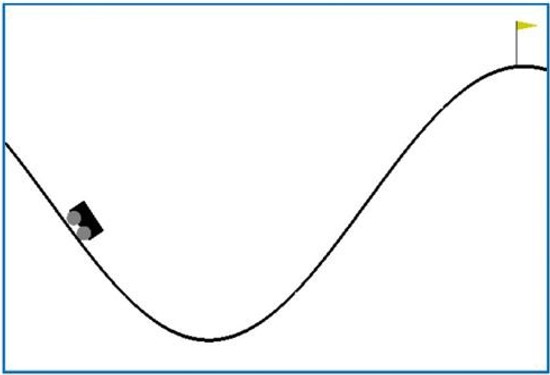}
\caption{\small The Mountain Car environment.}\label{Fig:4}
\end{figure}

In the second experiment, the Mountain Car platform, is shown in Fig.~\ref{Fig:4}, is chosen for the evaluation. Mountain Car is a classic RL problem where the objective is to create an algorithm which learns to climb a steep hill to reach the goal marked by a flag. The car's engine is not powerful enough to drive up the hill without a head start, therefore, the car must drive up the left hill to obtain enough momentum to scale the steeper hill to the right and reach the goal.
The state of the system is the position of the car and its velocity (i.e. $\s=[x, \dot{x}]^T$).  The possible actions are restricted within ($\mA =\{0, 1,2 \}$) which are ``push left'', ``no push'', and the ``push right'', respectively. The road ends at the position $-1.2m$, i.e., the position must be greater that $-1.2m$. The task is to reach the top  where the position must be larger or equal to $0.5m$. There would be a $-1$ reward for each step where the car is unable to mount the hill to reach the goal and  there is no penalty for climbing the left hill acting as a wall when is reached. Each episode starts with a random position ranging from $-0.6m$ to $-0.4m$ with no velocity.

Similar to the Inverted Pendulum experiment presented in Section~\ref{sec:cart_pole}, $9$ RBFs and a bias parameter are used for the proposed $\MMK$ algorithm. Therefore, the size of the feature vector is $30$. The mean and covariance of the RBFs are initialized as follows
\begin{eqnarray}
\!\!\!\!\!\!\u_{n,a_d}\!\! &\!\!\!\!\!\!\in\!\!\!\!\!\!&\!\!\!\! \{-0.775, -0.35, +0.775 \}\times \{-0.035, 0, +0.035\}\nonumber\!\!\!\!\!\!\\
\\
\!\!\!\!\!\!\Sig_{n,a_d} \!\!&\!\!\!\!\!\!=\!\!\!\!\!\!& \I_2.
\end{eqnarray}
The initial values of $\lambda_{\u}$ and $\lambda_{\Sig}$ are selected as $100$ and $80$, respectively. The time step is selected to be $50$ milliseconds and the discount factor is chosen as $0.95$. The process noise covariance is set to $\Q\k = 10^{-3}\I_{30}$, and the transition matrix is selected to be the identity matrix ($\F = \I_{30}$). Candidate values for $R$ are selected from the same set was selected for the Inverted Pendulum environment.
\begin{figure}[t!]
\centering
\includegraphics[scale=0.55]{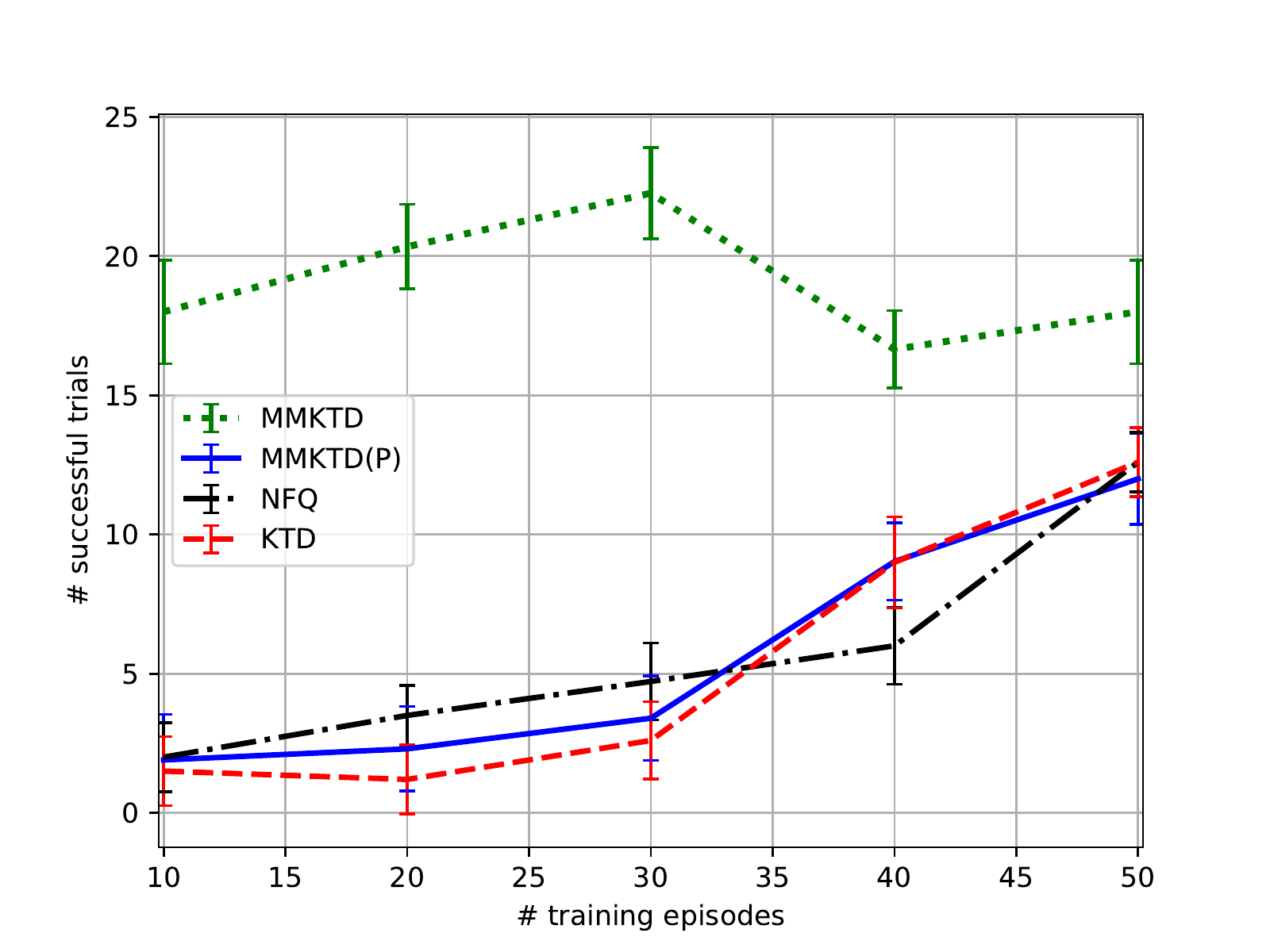}
\caption{\small{Similar to Fig.~\ref{Fig:2} but for the Mountain Car environment.}}\label{Fig:5}
\end{figure}

\begin{figure}[t!]
\centering
\includegraphics[scale=0.55]{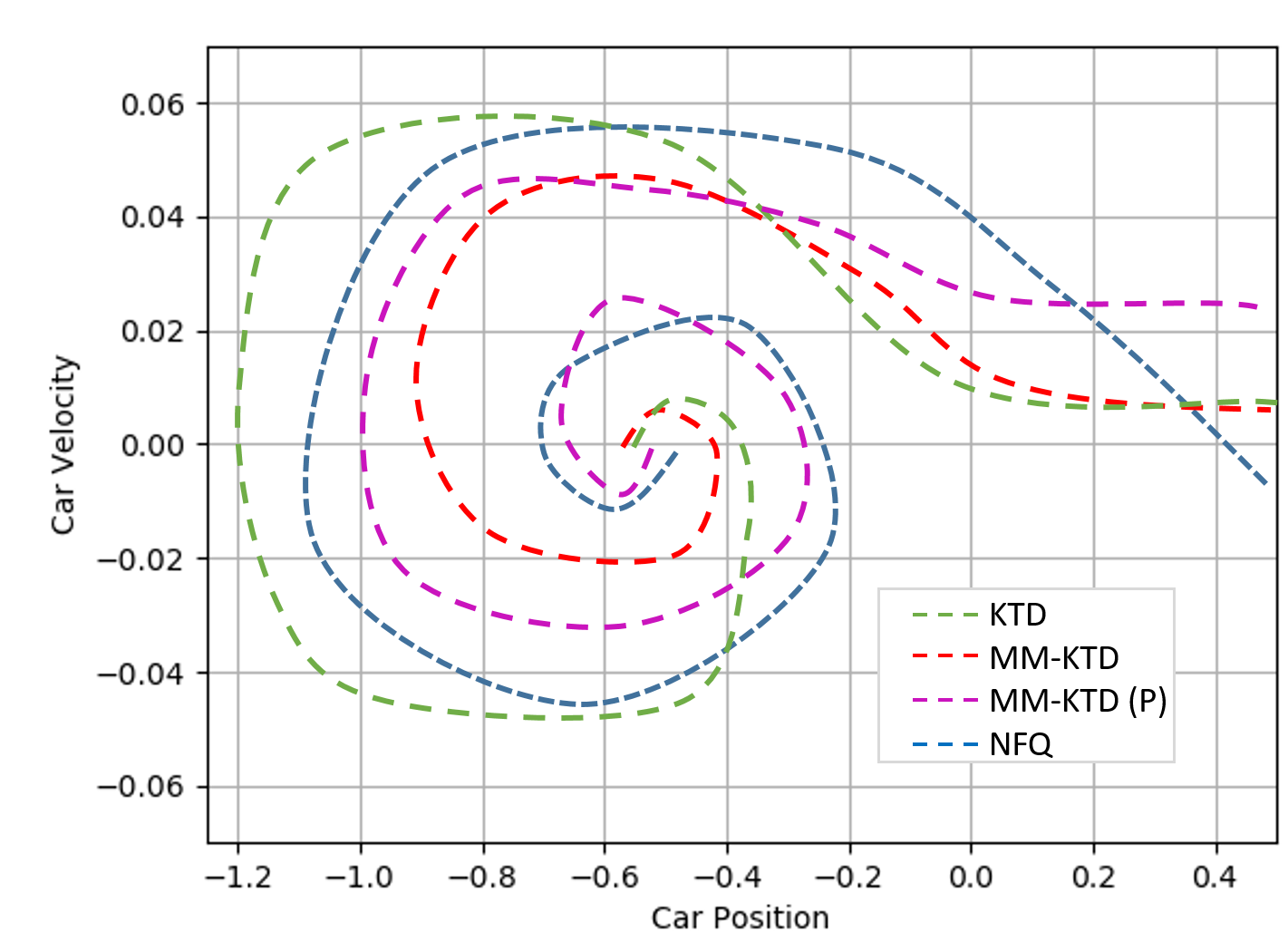}
\caption{\small State space trajectory of the Mountain Car environment at episode number 50.}\label{Fig:6}
\end{figure}

\begin{figure}[t!]
\centering
\includegraphics[scale=.7]{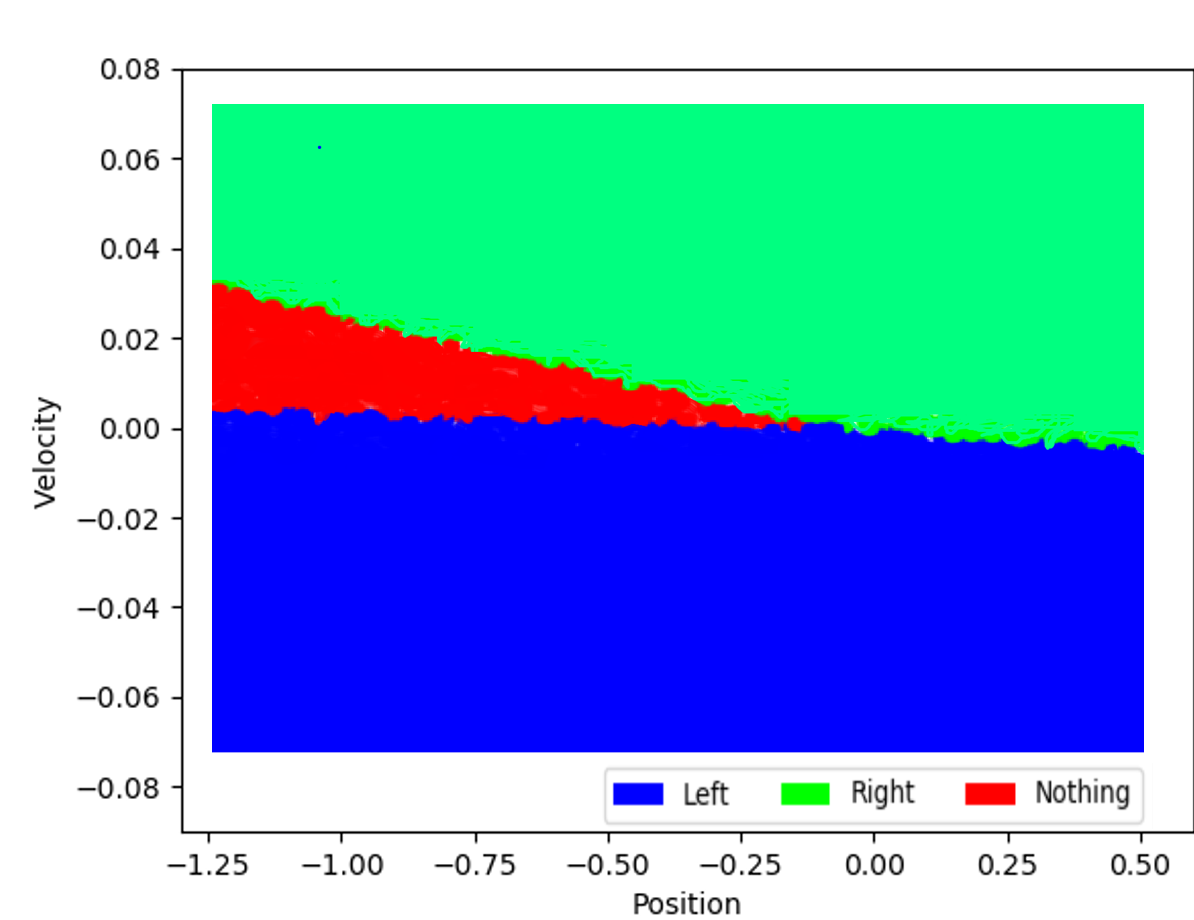}
\caption{\small Actions resulted from optimal policy of MM-KTD algorithm at the episode number 50 for Mountain Car environment.}\label{Fig:7}
\end{figure}

The initial weights are selected to be zero (i.e., $\bt_0 = \bm{0}_{30}$), whereas the initial error covariance is chosen as $\P_{\bt,0} = 10\I_{30}$. The agent first is trained through different number of episodes, then tested $10$ times for $50$ episodes to find if it can reach the flag during $200$ samples of each episode. Fig.~\ref{Fig:5} shows the results of KTD~\cite{23}, $\MMK$, $\MMK$ (P)), and NFQ learning method of~\cite{6}. As expected, the performance of KTD, $\MMK$ (P) and NFQ improves with increased training episodes.  However, KTD, $\MMK$ (P) and NFQ can't provide that level of performance is achieved by $\MMK$ with the lowest number of training episodes in this experiment ($10$). Based on the achieved results shown in Table~\ref{Table:2}, $\MMK$ is the most sample efficient approach of all tested algorithms.
Fig.~\ref{Fig:6} depicts the trajectories of the system's states (position, velocity) for the Mountain Car environment at episode number 50. As can be observed from  Fig.~\ref{Fig:6}, the episode starts from zero velocity and a random position in the range [$-0.6, -0.4$], and ends when the position reaches to $0.5m$. Fig.~\ref{Fig:7} shows the optimal agent's actions resulted from applying the MM-KTD scheme over a combination of positions and velocities. Based on Fig.~\ref{Fig:7}, the agent moves left when its velocity is negative. Most of the times that the velocity is positive, the agent moves to the right and sometimes does nothing. 

%

\begin{table}[t]
\vspace{-.1in}
\caption{\small The number (percentage) of successful trials of the proposed MM-KTD scheme as compared with other RL methods for the Mountain Car environment.}\label{Table:2}
\centering
\begin{tabular}{|c|p{1cm}|p{1.6cm}|p{1.1cm}|p{1.0cm}|}
\hline
\multirow{2}[2]{*}{\textbf{\!\!\!\# Training Episodes\!\!\!}} & \multirow{2}[2]{*}{\textbf{KTD}} &  \multirow{2}[2]{*}{\textbf{$\MMK$ (P)}} &  \multirow{2}[2]{*}{\textbf{$\MMK$}}&  \multirow{2}[2]{*}{\textbf{NFQ}} \\
&&&&\\
\hline
\!\!\!\multirow{2}[2]{*}{\textbf{10}}\!\!\!  & 1.5/50 (3\%) &  1.9/50 \,\,\,\,(3.8\%) &  18/50 (36\%) &  2.0/50 (4\%) \\
\hline
\!\!\!\multirow{2}[2]{*}{\textbf{20}}\!\!\!  & 1.2/50 (2.4\%) &  2.3/50 \,\,\,\,(4.6\%) &  20.3/50 (40.6\%) &  3.5/50 (7\%) \\
\hline
\!\!\!\multirow{2}[2]{*}{\textbf{30}}\!\!\!  & 2.6/50 (5.2\%) &   3.4/50 \,\,\,\,(6.8\%) &  22.2/50 (44.4\%) & 4.72/50 (9.44\%)\\
\hline
\!\!\!\multirow{2}[2]{*}{\textbf{40}}\!\!\!  & 9.2/50 (18.4\%) &  9/50 \,\,\,\,\,\,\,\,(18\%) &  16.6/50 (33.2\%) & 6/50 (12\%) \\
\hline
\!\!\!\multirow{2}[2]{*}{\textbf{50}}\!\!\!  & 12.3/50 (24.6\%) &   12.0/50 (24\%) &  18.0/50 (36\%) & 12.6/50 (25.2\%)\\
\hline
\end{tabular}
\end{table}
\subsection{Lunar Lander}  \label{sec:lunar_lander}
\begin{figure}[t!]
\centering
\includegraphics[scale=0.45]{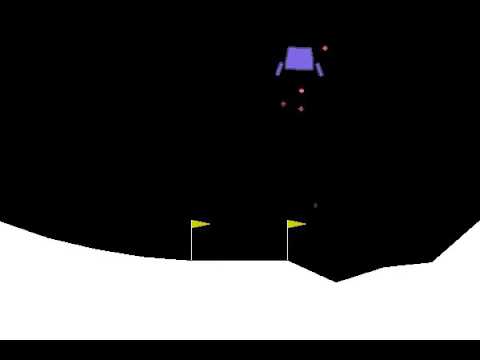}
\caption{\small{The Lunar Lander environment.}}\label{Fig:8}
\end{figure}

\begin{figure}[t!]
\centering
\includegraphics[scale=0.55]{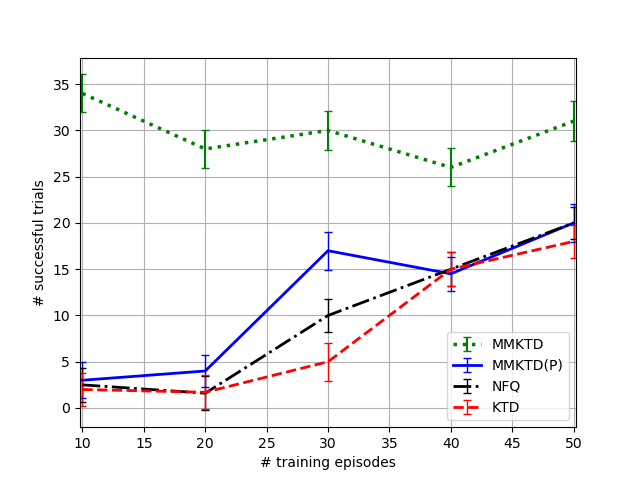}
\caption{\small{Similar to Fig.~\ref{Fig:2} but for the Lunar Lander environment.}}\label{Fig:9}
\end{figure}

In a third experiment, we focus on the Lunar Lander environment, which is a more complicated environment compared to the two previous ones. In the  Lunar Lander environment, the goal is for the RL agent to learn to land successfully on a landing pad located at coordinate  $(0,0)$ in a randomly generated surface on the moon as shown in Fig.~\ref{Fig:8}. The state space of the system consists of the agent's position $(x,y)$ in space, horizontal and vertical velocity $(v_x,v_y)$, orientation in space $\theta$, and angular velocity $\dot{\theta}$. The agent has four possible actions, i.e., do nothing; firing the left engine, firing the main engine, and; firing the right engine ($\mA =\{0, 1, 2, 3 \}$).  Reward for landing on the pad is about $100$ to $140$ points, varying on the lander placement on the pad. If the lander moves away from the landing pad it loses reward. Each episode terminates if the lander lands or crashes, receiving additional +$100$ or -$100$ points, respectively. Each leg ground contact worth +$10$ points. Firing the main engine results in a -$0.3$ point penalty for each frame. The problem is considered solved if the agent receives +$200$ points over $100$ iterations. The RBFs of order two are considered for each state variable resulting in $64$ RBFs for each action. Consequently, the size of the feature vector $\bm{\phi}(\s\k,a\k)$ will be $256$. Based on the expected range of each variable of the state vector, the initial mean and covariance of the RBFs are chosen as follows
\begin{eqnarray}
\u_{n,a_d} &\in& \{-0.333, +0.333\}^6,
\\
\Sig_{n,a_d}&=& 2\I_{6}.
\end{eqnarray}
The initial values of $\lambda_{\u}$ and $\lambda_{\Sig}$ are both selected as $200$ to keep the system stable. The discount factor is selected as $0.99$. The process noise covariance is set to $\Q\k=10^{-1}\I_{256}$. Candidate values for $R$ are selected from the same set as was used for the Inverted Pendulum environment. Like the two previous environments, first, the agent is trained through different number of episodes changing from $10$ to $50$, then tested $10$ times for $50$ trials to find if the agent can successfully land on the landing pad over $100$ steps. Fig.~\ref{Fig:9} depicts the results of applying the KTD~\cite{23}, $\MMK$, $\MMK$ (P)), and NFQ learning~\cite{6} scheme. It can be observed that the proposed approach outperforms its counterparts.
\subsection{Parameters Selection}  \label{sec:parameter_selection}
\vspace{.1in}
\noindent
In this sub-section, we provide more details on different values assigned to the underlying parameters in the three simulations presented above. First, it is worth mentioning that one aspect of the proposed multiple-model type framework is to address the issues of reducing dependence of RL performance on its parameter values. For adaptation of the RBFs' parameters, the initial centers of the RBFs are, typically, distributed evenly along each dimension of the state vector, leading to $n^d$ centers for $d$ state variables and a given order $n$. Initial variance of each state's variable $(\sigma^2)$ is often set to $\frac{2}{n-1}$. For the Inverted Pendulum and Mountain Car, the value of $n = 3$ is selected following the KTD approach of~\cite{23} for having fair comparison. For the Lunar Lander, the value of $n$ is selected to be $2$. Furthermore, the values for $\lambda_{\bm{\mu}}$ and $\lambda_{\Sig}$ are selected in such a way to keep the system stable. In addition, based on Fig.~\ref{Fig:stab}, it can be observed that RBFs are fairly stable against selection of their underlying parameters. The discount factor denoted by $\gamma$ affects how much weight is given to the future rewards in the value function. A discount factor $\gamma=0$ will result in state/action values representing the immediate reward, while a higher discount factor will result in the values representing the cumulative discounted future reward that an agent is expected to receive (behaving under a given policy). Commonly (as is the case in the  implemented environments), a large portion of the reward is earned upon reaching the goal. To prioritize this final success, we expect an acceptable $\gamma$ to be close 1. In our manuscript, $\gamma$ is set to the high values of $0.95$ and $0.99$.

\begin{figure}[t!]
\centering
\includegraphics[scale=0.4]{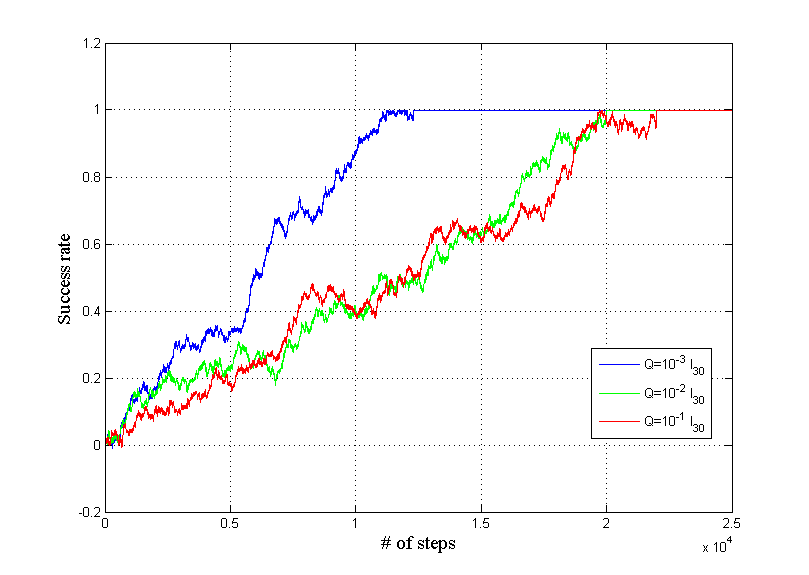}
\caption{\small Success rate during 50 training episodes for Inverted Pendulum environment.}\label{Fig:sens}
\end{figure}

The prior $\bm{\theta}_0$ should be initialized to a value close to the expected optimal value based on historical data, or to a default value, e.g., the zero vector. The prior $\bm{P}_0$ quantifies the certainty that the user has in the initialized prior $\bm{\theta}_0$, the lower the less certainty. The process noise covariance is a design parameter, the proposed MM-KTD algorithm allows, systematically, to use a set of different values of process noise covariance ($\Q$).  We have conducted a sensitivity analysis experiment for the Inverted Pendulum environment to evaluate sensitivity to this design parameter for the scenario where only a single initial value can be assigned. Fig.~\ref{Fig:sens} presents the agent's success rate during the training phase over $50$ episodes based on three different assignments to $\Q$. It can be observed that different values of the process noise covariance affect the time that the agent takes to complete the training process. Finally and as stated previously, variable $R$, i.e., the measurement noise variance, is one of the most important parameters to be identified. To select this parameter, our intuition in the proposed MM-KTD framework is to cover the potential range of the measurement noise variance using $M$ mode-matched filters. The parameter $M$ is set a-priori denoting the number of candidates $R\i$, for ($1 \leq i \leq M$), for the observation noise variance. For example, in the experiments, we have selected $M$ to be equal to $11$.
\section{Conclusion}  \label{sec:con}
The paper proposes a Multiple-Model Kalman-filter-based Temporal Differences framework, referred to as the $\MMK$, which deals with the problems of sample efficiency, online learning, prior information and memory problems in the other RL algorithms. The proposed $\MMK$ algorithm is evaluated based on three RL experiments. Based on the achieved results, the proposed algorithm achieved the highest number of successful trials while its number of training episodes for learning the best policy is considerably less in comparison to its counterparts. The need for restricted number of learning episodes results in the reduced training data, which is critical for practical applications in the real scenarios.

\bibliographystyle{IEEEbib}

\begin{IEEEbiography}[{\includegraphics[width=1in,height=1.0in,clip,keepaspectratio]{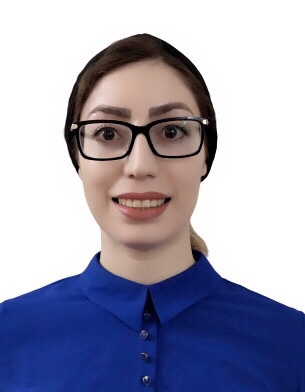}}]{Parvin Malekzadeh}, received the B.Sc. degree in electrical engineering from Sharif University of Technology, Iran, in 2018. She is currently pursuing the M.Sc. degree at Concordia University, Canada. Her research interests lie in the areas of signal processing, Machine Learning, Reinforcement Learning and Internet of Things, with a focus on indoor localization.
\end{IEEEbiography}
\begin{IEEEbiography}[{\includegraphics[width=1in,height=1.25in,clip,keepaspectratio]{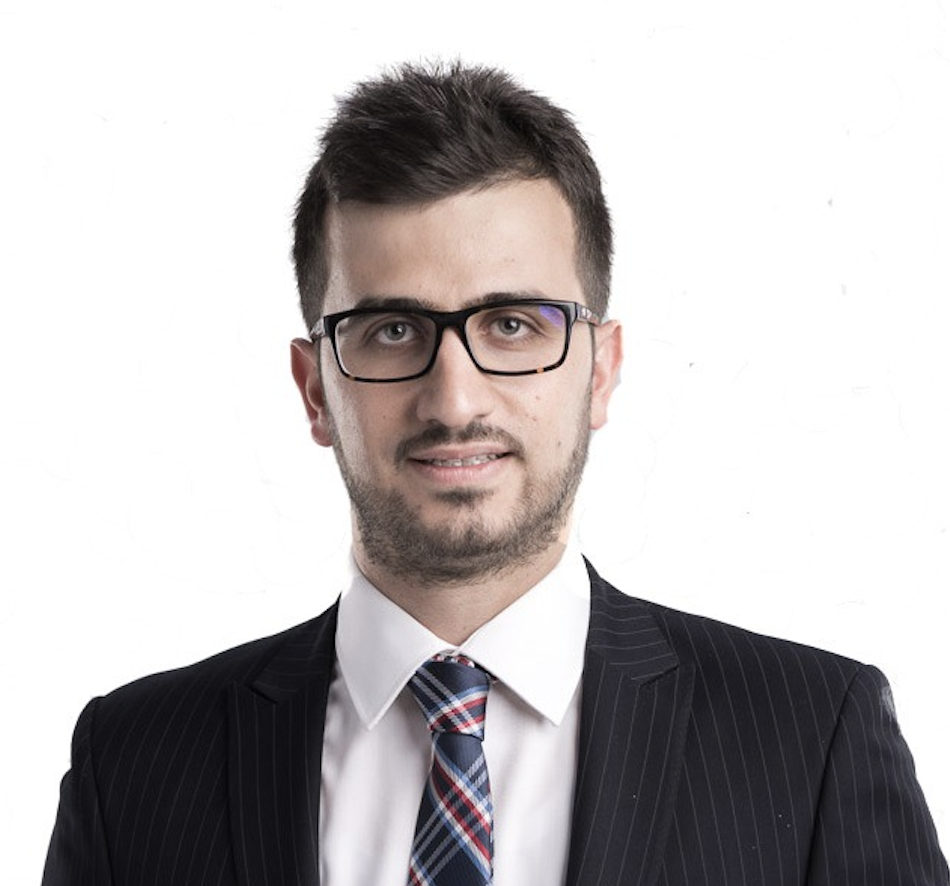}}]{Mohammad Salimibeni}, received the B.Sc. degree in Electrical Engineering from the Isfahan University of Technology (IUT), in 2011 and the M.Sc. degree in Electrical Engineering from the same university (IUT) in 2014. He is currently pursuing the Ph.D. degree  at Concordia University. His research interests include signal processing, Reinforcement Learning and distributed IoT-based networks with a particular interest in distributed ledger technology and Blockchain.
\end{IEEEbiography}

\begin{IEEEbiography}[{\includegraphics[width=1in,height=1.25in,clip,keepaspectratio]{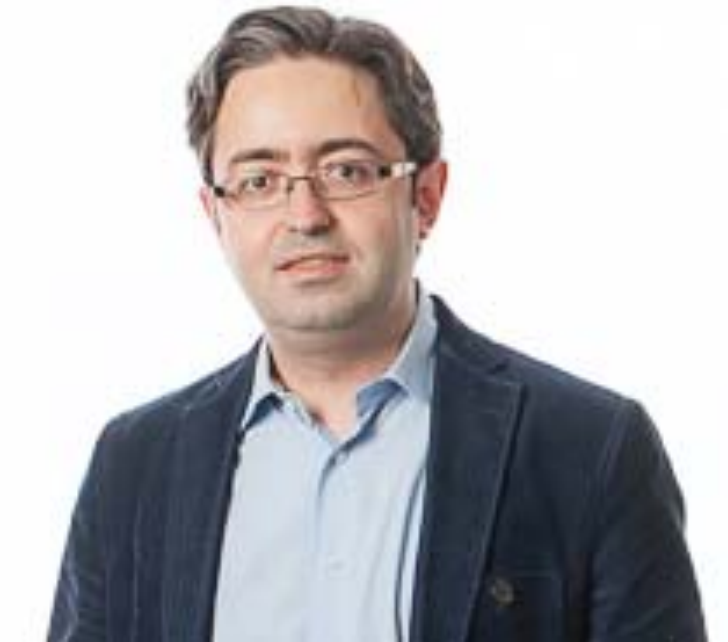}}]{Arash Mohammadi}
	(S'08-M'14-SM'17) is an Assistant Professor with Concordia Institute for Information Systems Engineering (CIISE), Concordia University, Montreal, Canada. He received B.Sc. degree from University of Tehran in 2005, M.Sc. degree from Amirkabir University of Technology (Tehran Polytechnic) in 2007, and Ph.D. from York University in 2013. Prior to joining Concordia University, he was a Postdoctoral Fellow at Department of Electrical and Computer Engineering, University of Toronto, Canada. He is a registered Professional Engineer in Ontario. He is the Director-Membership Services  of IEEE Signal Processing Society (SPS). His research interests include: cyber-physical systems; information fusion; distributed signal processing for agent networks; secure networked control systems; biomedical signal processing; consensus algorithms, large-scale dynamical systems, and; smart grids. Dr. Mohammadi has received several distinguishing awards, including the Eshrat Arjomandi Award for outstanding Ph.D. dissertation from Electrical Engineering and Computer Science Department of York University in 2013, and Concordia President's Excellence  in Teaching Awards in 2018.
\end{IEEEbiography}

\begin{IEEEbiography}[{\includegraphics[width=1in,height=1.25in,clip,keepaspectratio]{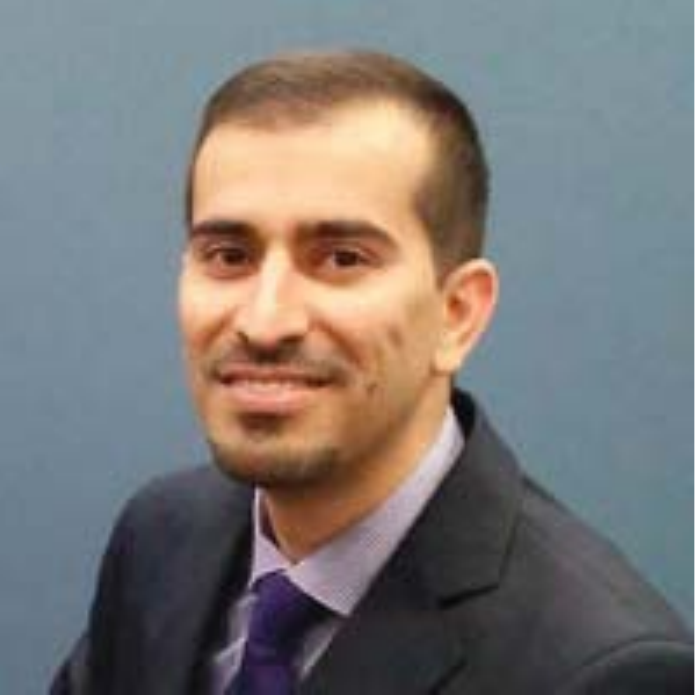}}]{Akbar Assa} (S'08–M'15) received the Ph.D. degree in Mechanical Engineering from Ryerson University in 2015. He was a Post-Doctoral Fellow with Ryerson University in 2015 and served as a Professor with the Sheridan College from 2015 to 2016. He joined the Multimedia Laboratory, ECE Department, University of Toronto, in 2017, as a
NSERC Post-Doctoral Fellow. His research interests span from robotics, visual servoing, and
control systems to image processing, optimal estimation, and sensor fusion.
\end{IEEEbiography}

\begin{IEEEbiography}[{\includegraphics[width=1in,height=1.25in,clip,keepaspectratio]{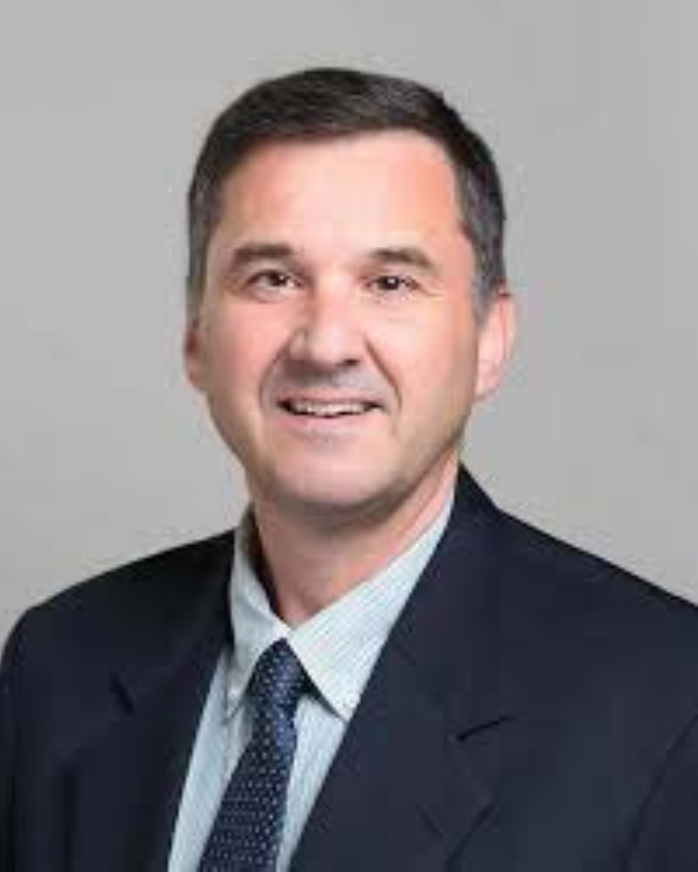}}]{Konstantinos N.  Plataniotis}(S'93-M'95-SM'03-F'12)  is currently a Professor and the Bell Canada Chair in multimedia with the ECE Department, University of Toronto. He is also the Founder and the Inaugural Director-Research for the Identity, Privacy and Security Institute (IPSI), University of Toronto, where he was the Director for the Knowledge Media
Design Institute (KMDI), from January 2010 to July 2012. He is a registered Professional Engineer in Ontario. Among his publications in these fields are the recent books WLAN Positioning Systems (2012) and Multi-Linear Subspace Learning: Reduction of Multidimensional Data (2013). His research interests are knowledge and digital media design,multimedia systems, biometrics, image and signal processing, communications systems, and pattern recognition. Dr. Plataniotis is a fellow of the Engineering Institute of Canada. He has
served as the Editor-in-Chief for the IEEE SIGNAL PROCESSING LETTERS, and as the Technical Co-Chair for the IEEE 2013 International Conference in Acoustics, Speech and Signal Processing. He was the Vice President of the IEEE Signal Processing Society for Membership from 2014 to 2016. He was the General Co-Chair of the 2017 IEEE GlobalSIP, the 2018 IEEE International Conference on Image Processing (ICIP 2018), and is the General Co-Chair of the 2021 IEEE International Conference on Acoustics, Speech and Signal Processing (ICASSP 2021).
\end{IEEEbiography}

\EOD

\end{document}